\documentclass{article}

\PassOptionsToPackage{numbers, compress}{natbib}

\usepackage[preprint]{neurips_2023}

\usepackage[utf8]{inputenc} 
\usepackage[T1]{fontenc}    
\usepackage{hyperref} 
\usepackage{url}   
\usepackage{amsfonts} 
\usepackage{amsmath}
\usepackage{nicefrac} 
\usepackage{microtype}  

\usepackage{graphicx}
\usepackage{arydshln}
\usepackage{booktabs}
\usepackage{multirow}
\usepackage{makecell}
\usepackage{graphicx}
\usepackage{subfigure}
\usepackage{wrapfig}
\usepackage{makecell}
\usepackage[dvipsnames]{xcolor}
\usepackage{float}
\usepackage{tikz}

\newcommand{\nameofmethod}{PLLaVA }

\hypersetup{
    colorlinks=true,
    linkcolor=red,
    citecolor=cyan,
    filecolor=magenta,      
    urlcolor=cyan,
    }

\title{\nameofmethod: Parameter-free LLaVA Extension from Images to Videos for Video Dense Captioning  }
\author{%
Lin Xu$^{1*}$, Yilin Zhao$^{2*}$, Daquan Zhou$^{3\star\dagger}$, Zhijie Lin$^3$, See Kiong Ng$^1$, Jiashi Feng$^3$ \\
[0.2cm]
$^1$National University of Singapore \quad 
$^2$New York University \quad
$^3$Bytedance 
}

\newcommand{\greentext}[1]{\color{ForestGreen}{#1}}
\newcommand{\redtext}[1]{\color{red}{#1}}

\makeatletter
\def\blfootnote{\xdef\@thefnmark{}\@footnotetext}
\makeatother

\begin{document}
\maketitle

\begin{tikzpicture}[remember picture,overlay,shift={(current page.north east)}]
    \node[anchor=north east,xshift=-2.9cm,yshift=-3.9cm]{\includegraphics[width=2cm]{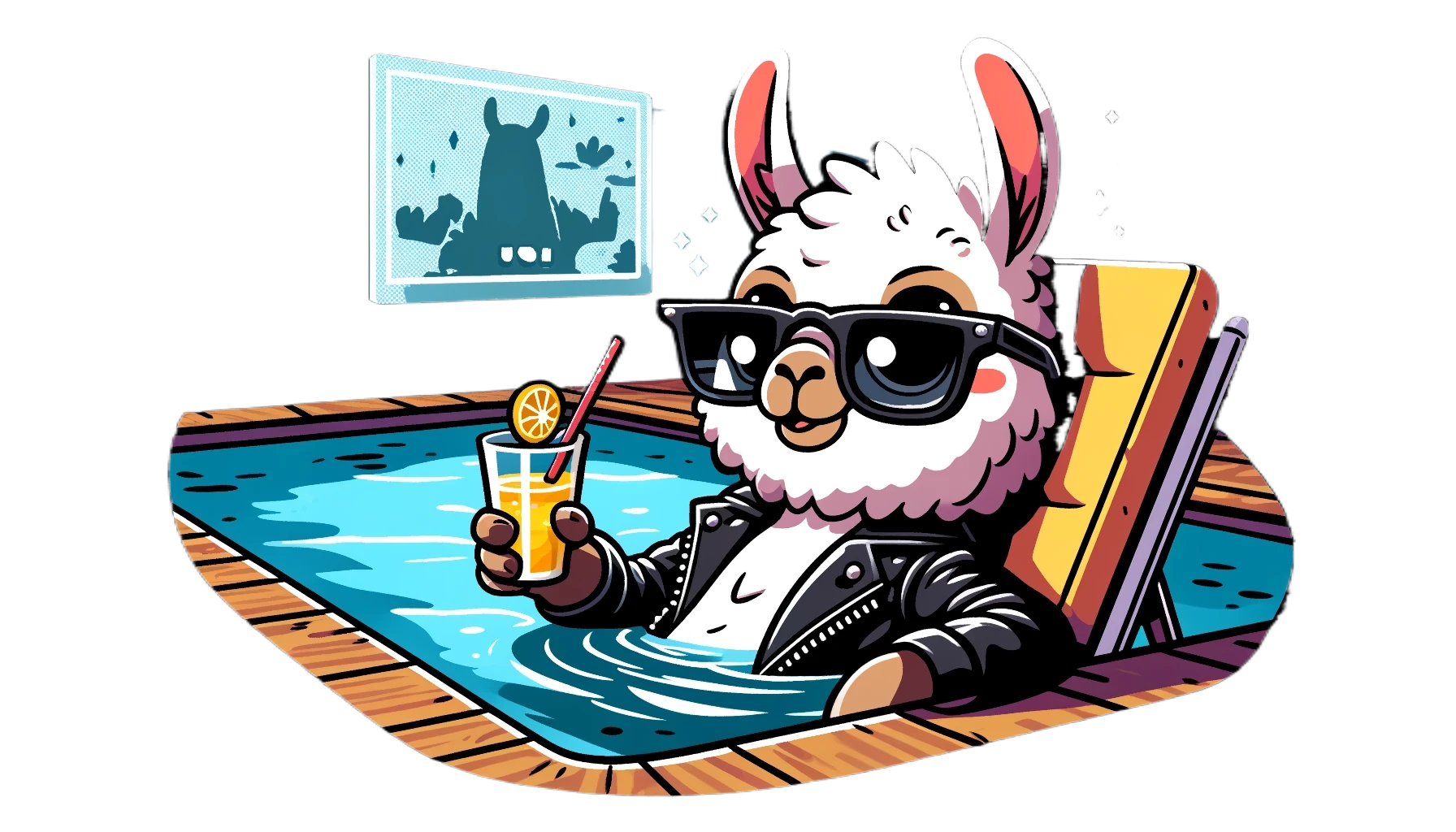}};
\end{tikzpicture}




\blfootnote{$^{*}$ Equal contribution. $^{\star}$Project lead. $^{\dagger}$Corresponding authors. Lin Xu, cathyxl2016@gmail.com; Daquan Zhou, zhoudaquan21@gmail.com}

\vspace{-9mm}
\begin{figure}[H]
\vspace{-15pt}
\centering
\subfigure[
\nameofmethod generates dense descriptions of the video contents including motions, and attires. ]{
    \begin{minipage}[b]{0.93\textwidth}
        \includegraphics[width=1\textwidth]{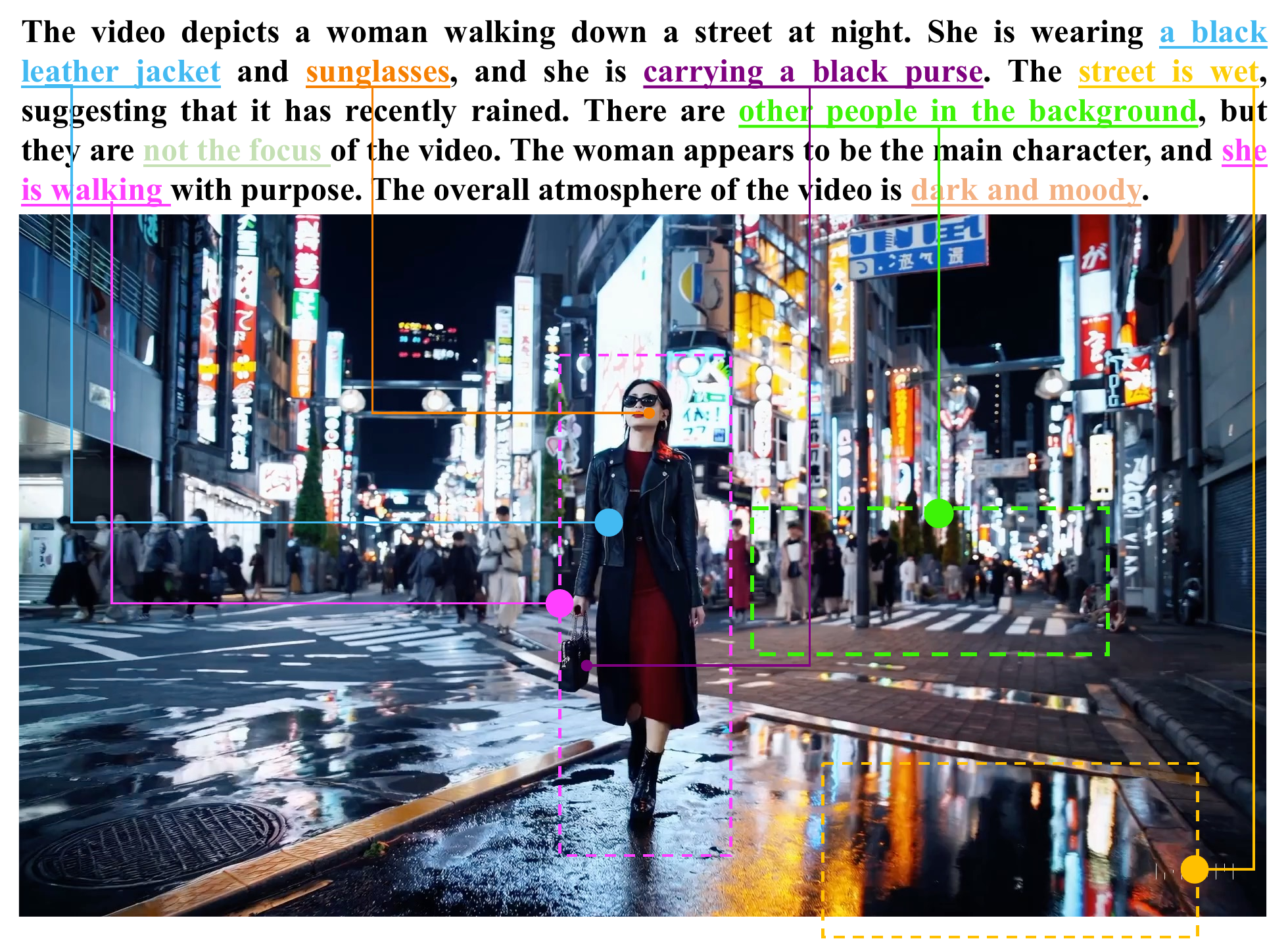}
    \end{minipage}
    \label{subfig:qualitative}
    } \\

\subfigure[State-Of-The-Arts on various video understanding tasks.
]{
    \begin{minipage}[b]{0.56\textwidth}
        \includegraphics[width=1\textwidth]{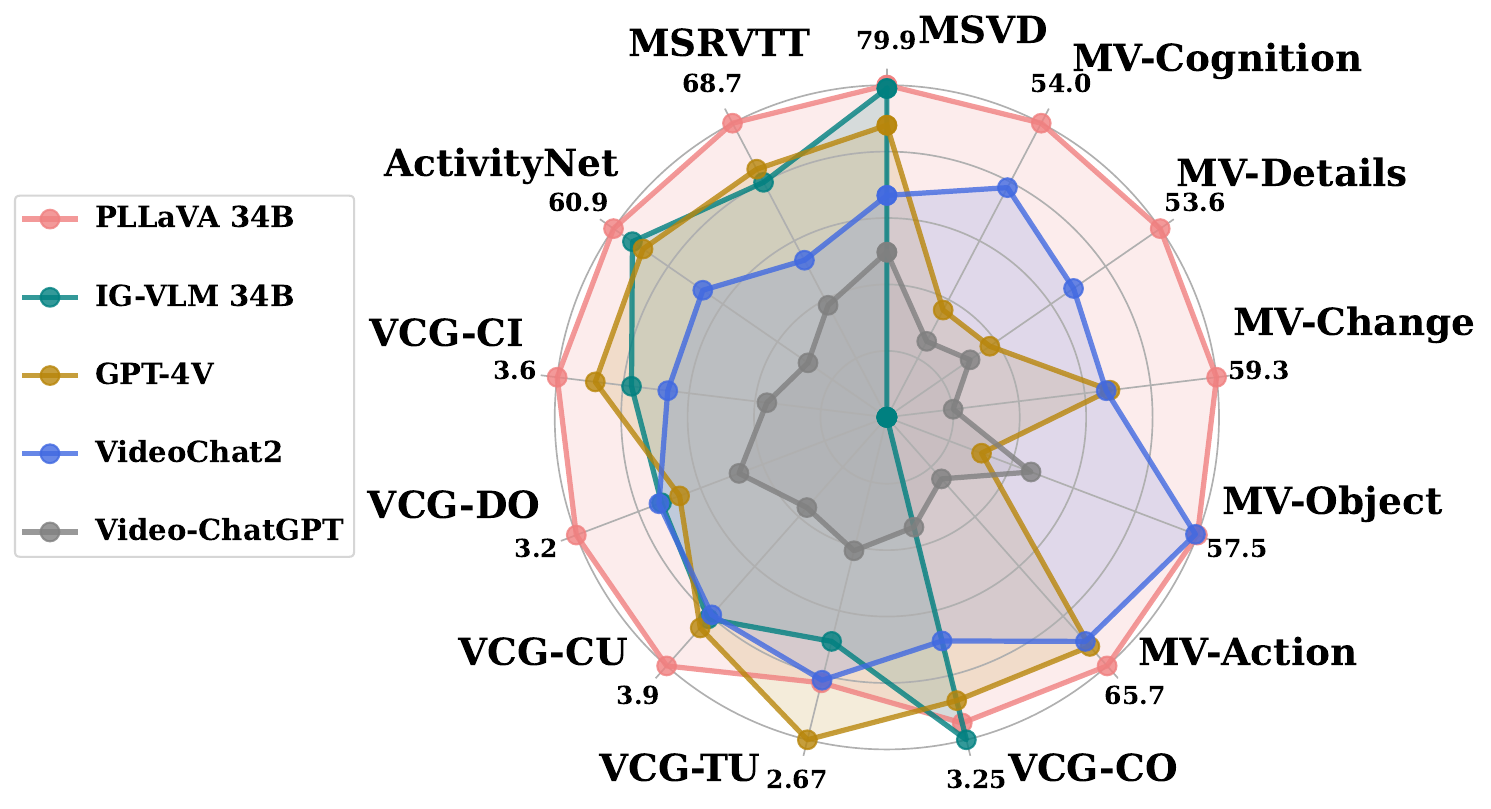}
    \end{minipage}
    \label{subfig:radar}
    }
\subfigure[Better model scaling performance.
]{
    \begin{minipage}[b]{0.35\textwidth}
	\includegraphics[width=1\textwidth]{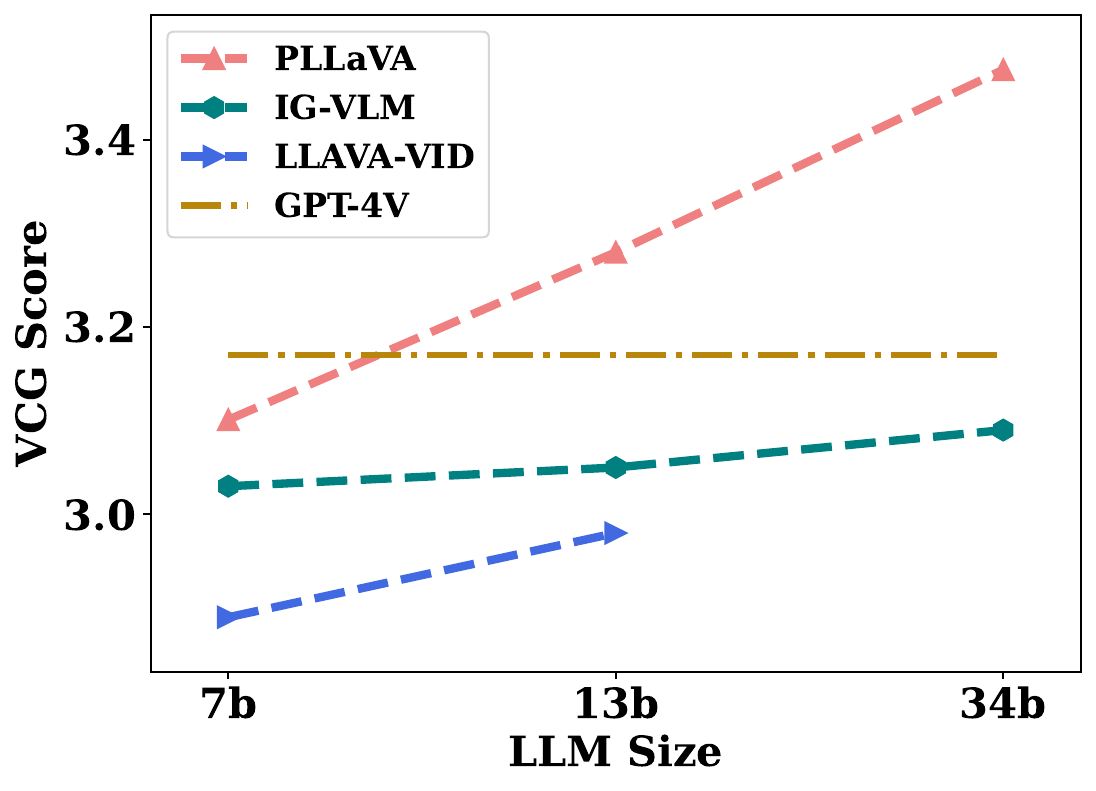}
    \end{minipage}
    \label{subfig:llm_scaling_with_model}
 } 
 \vspace{-5pt}

\caption{Performance presentation of \nameofmethod. (a) An example of captions generated with \nameofmethod 34B. (b) Performance comparison of \nameofmethod with recent strong baselines over different video benchmarks and (c) the scaling curve of \nameofmethod and recent SOTA methods.
}
\label{fig:teaser}
\vspace{-5pt}
    
\end{figure}


\begin{abstract}
  Vision-language pre-training has significantly elevated performance across a wide range of image-language applications. Yet, the pre-training process for video-related tasks demands exceptionally large computational and data resources, which hinders the progress of video-language models.
  This paper investigates a straightforward, highly efficient, and resource-light approach to adapting an existing image-language pre-trained model for dense video understanding. 
  Our preliminary experiments reveal that directly fine-tuning pre-trained image-language models with multiple frames as inputs on video datasets leads to performance saturation or even a drop. Our further investigation reveals that it is largely attributed to the bias of learned high-norm visual features.  
  Motivated by this finding, we propose a simple but effective pooling strategy to smooth the feature distribution along the temporal dimension and thus reduce the dominant impacts from the extreme features. The new model is termed Pooling LLaVA, or \nameofmethod in short.  \nameofmethod achieves new state-of-the-art performance on modern benchmark datasets for both video question-answer and captioning tasks. Notably, on the recent popular Video ChatGPT benchmark, PLLaVA achieves a score of 3.48 out of 5 on average of five evaluated dimensions, exceeding the previous SOTA results from GPT4V (IG-VLM) by 9\%. On the latest multi-choice benchmark MVBench, PLLaVA achieves 58.1\% accuracy on average across 20 sub-tasks, 14.5\% higher than GPT4V (IG-VLM). Code is available at \url{https://pllava.github.io}. 
  %
\end{abstract}

\section{Introduction}

Multimodal Large Language Models (MLLMs) have demonstrated remarkable proficiency in image comprehension when trained on large-scale image-text pairs~\cite{li2023blip2,zhu2023minigpt,liu2024visual,liu2023improved,huang2024lita}. 
Analogous to the image domain, the recent video understanding models also explore a similar pipeline to fine-tune LLMs on large-scale video-text data~\cite{chen2023videollm, li2023videochat, li2023mvbench}. However, this method suffers a high cost of computing resources and video data annotations. A more pragmatic approach is to \emph{adapt} the pre-trained image-domain MLLMs to video data~\cite{maaz2023videochatgpt, liu2023btadapter, kim2024igvlm}. 

An intuitive method for image MLLM adaption is to encode multiple video frames into a sequence of features and directly feed them into MLLMs, as the Large language Models(LLMs)~\cite{vaswani2017attention,touvron2023llama} are native for processing sequential features and shown to be capable of understanding temporal information~\cite{lian2023llm, liu2024st}. However, we empirically found two technical challenges when extending image MLLMs to video data in this way. First, compared to zero-shot applications, training the image MLLM on video data does not always increase the performance but introduces performance vulnerability to the change of inquiry prompts. Secondly, increasing the size of the language model component does not improve the video understanding performance. Those two observations are counter-intuitive since scaling up model sizes and exposing models to more downstream data are typically considered beneficial for model performance. 

\begin{figure}[t]
    \centering
    \includegraphics[width=0.98\linewidth]{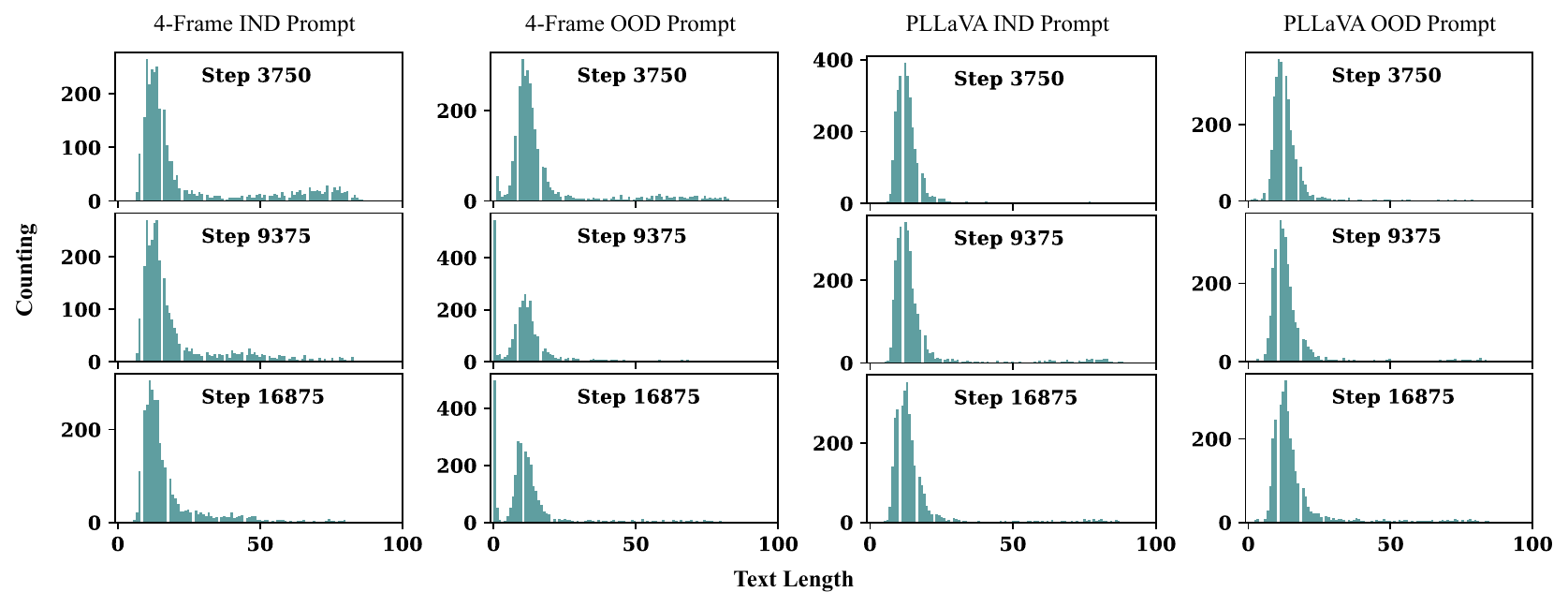}
    \caption{
    Histograms of generation text length distribution for 4-Frame and \nameofmethod. x-axis are text lengths, y-axis indicates the counting of text-lengths. 4-Frame generates shorter texts under more training steps and the out-of-distribution prompt, while \nameofmethod remains consistent in both situations. 
    } 
    \label{fig:text_length_compare}
\end{figure}

We then conducted a series of studies to investigate the root cause of these two observations. 
For the first one, we found it is mainly due to the limited information encoded by the image encoder. 
When experimenting on LLaVA~\cite{liu2024visual} with 4-frame inputs, we experimentally found that, as shown in~\autoref{fig:norm_hist_w_gen}, some visual feature tokens have dominantly larger norms over the others during the fine-tuning process. 
These tokens lead to shorter text descriptions with lower quality. As demonstrated in~\autoref{fig:text_length_compare}, the 4-frame models tend to generate shorter texts with training on more samples.  We conjecture that the large-norm features have obtained global video information and thus suppress the norms of other tokens,  due to the softmax calculation during the self-attention. This leads the generated description to be short. Even worse, if the prompt template changes, the learned MLLMs would completely collapse, leading to rather short descriptions or even no response. 
We observe that adding more video frames could mitigate the suppression of the majority of the tokens. However, this would lead to significantly larger memory consumption.

Thus, there is a trade-off between the number of frames and the computation cost. 
%
The intuitive way is to downsample the video frames.
However, directly averaging the spatial and temporal dimensions as has been done in VideoChatGPT \cite{maaz2023videochatgpt} loses too much spatial information and also does not achieve optimal performance during the scaling of the training dataset. Thus, the target is to find the minimum spatial resolution of each frame that does not degrade the scaling curve.
To achieve this, we adopt a pooling~\cite{lecun1989handwritten} operation to explore the optimal settings such that it does not degrade the benefits of increasing the temporal receptive field. The impact of the pooling operation is shown in Figure~\ref{fig:pooling_shape}.

For the second observed phenomenon, we believe one main reason is the poor quality of the video dataset, compared to the image dataset. 
Specifically, many of the video datasets are in question-answer formats and the descriptions of the videos might be short. Thus, as the model learns the temporal description from the video dataset, the description of other metrics such as the objects and the spatial relations degrades. the stronger the LLM is, the faster the output degrades. 
%
Instead of building high-quality video datasets, we choose to explore architectural and optimization algorithms to better preserve the learned information in image datasets during the learning of the temporal information on video datasets. To achieve this, we utilize the tricks of weight fusion. We set two groups of weights: one from the image pre-raining and one with video dataset fine-tuning. After training, we searched to find the optimal combination of the image-based model weights and the video-based model weights in the hope that the combined model could gather benefits from both datasets. The process is termed post-training optimization in this paper and its impacts are shown in Figure \ref{fig:scaling_curves}. 

\begin{itemize}

    \item We performed a thorough initial investigation for directly applying image large multi-modality models to video tasks and found several failure modes. We then introduce an elegantly simple yet highly potent pooling strategy that systematically achieves the optimal balance between training efficiency and caption accuracy.
    \item We introduce a post-training model merging method that could effectively reduce the forgetting phenomenon of the large language models during multi-modality fine-tuning. With this, we are able to get a large video multi-modality model with 34B LLMs without the extra creation of high-quality datasets.
    \item We conduct extensive experiments to verify the superiority of the proposed model and achieve a new state-of-the-art across various video understanding benchmarks, especially for video captioning tasks with dense captions. With Pool-LLaVA, we do the re-captioning of the top 1M video data from Panda-70M with highly dense and accurate bilingual captions.
\end{itemize}



\section{Related Works}
\paragraph{Video Multimodal Large Language Models}
Video Multi-modality Models process video input and generate responses according to user commands. Commonly, they incorporate a projection network~\cite{maaz2023videochatgpt,Lin2023VideoLLaVALU,Li2023LLaMAVIDAI}, inter-modality attention~\cite{li2023videochat,li2023mvbench} or a modality perceiver~\cite{zhang2023VideoLLAMA,Song2023MovieChatFD,Jin2023ChatUniViUV} as learnable interfaces. These interfaces are instrumental in melding the spatial-temporal dynamics of videos with large language models' (LLMs) processing capabilities \cite{touvron2023llama,roumeliotis2023llama,chiang2023vicuna}, by transforming video content into a sequence of tokens that LLMs can adeptly analyze.
Parameter efficient learning schemes \cite{hu2021lora,lian2022scaling, feng2024mixture} are adapted to reduce the computational cost. Among them, BLIP~\cite{li2023blip2} marked a significant milestone by integrating a frozen vision encoder with BLIP to enhance video processing efficiency, with only the newly added Q-Former learnable. Demonstrating remarkable zero-shot capabilities in Video Question Answering (VQA), it outperformed existing techniques of its time.
Extending the innovations of its predecessors, Video-ChatGPT~\cite{maaz2023videochatgpt} introduced the trailblazing approach of video instruction tuning, along with creating a dataset of high-quality instructional data. This initiative set a new standard for assessing models through video-based text generation benchmarks.
VideoChat~\cite{li2023videochat} employed cross-attention mechanisms to skillfully condense video tokens, aligning user queries with the dialogue context to enhance the model's interpretative capabilities.
Building upon these advances, VideoChat2~\cite{li2023mvbench} refined the approach with a multi-stage bootstrapping technique that honed in on modality alignment and instruction tuning, amassing a robust collection of high-quality video data for fine-tuning instruction-driven tasks. VILA~\cite{lin2023vila} proposes more advanced training recipes.
Further integrating modalities, Video-LLaVA~\cite{Lin2023VideoLLaVALU} leveraged a pre-aligned encoder adaptable to both images and videos, facilitating shared projections and enabling synergistic training across image and video-related tasks. CAT~\cite{ye2024cat} introduces both video and audio to futher enhance understanding.

Long videos present significant challenges due to their intrinsic high computational complexity and extensive memory requirements. Handling the entire span of a long video with video tokens poses difficulties in jointly capturing spatial details and temporal dynamics effectively.
In response, Video Language Models (Video MLLMs) have adopted sophisticated temporal modeling techniques to address these challenges with greater efficiency.
MovieChat~\cite{Song2023MovieChatFD} implemented a novel memory-based mechanism within transformers, strategically combining similar frames to reduce both computational load and memory footprint.
Chat-UniVi~\cite{Jin2023ChatUniViUV} debuted a harmonized approach for processing images and videos, innovatively condensing spatial and temporal tokens through dynamic token merging, utilizing k-NN algorithms for improved efficiency.
LLaMA-VID~\cite{Li2023LLaMAVIDAI} innovated with a dual-token approach that effectively condensed video representations by segregating context and content tokens, allowing for more efficient compression.
VTimeLLM~\cite{huang2023vtimellm} emphasize the boundaries of videos by introducing a new question answering dataset.
Advancing this innovation, Vista-LLaMA~\cite{Ma2023VistaLLaMARV} introduced EDVT-Attention along with a sequential vision projector that meticulously curates visual tokens and condenses temporal tokens, progressively amalgamating them with a Q-former mechanism.
To further optimize the handling of extended videos, certain models emphasized the selective processing of keyframes, thus diminishing the volume of video frames required and streamlining the overall computational demands.

\paragraph{Pipelined Video Understanding}
Capitalizing on the Video MLLM framework, a novel approach emerged involving the use of pre-existing Video Models coupled with LLMs through a multi-stage process of video modality conversion.
This method entails translating video content into textual narratives, typically through the employment of pretrained VideoLMs, before integrating with an LLM in the final phase. By encapsulating videos as text tokens, it leverages the LLMs' adeptness at navigating textual data, thereby permitting the interpretation of temporal sequences via these crafted descriptions.
VideoChat-Text~\cite{li2023videochat} adeptly converts video streams into comprehensive text descriptions, encapsulating a range of video elements. Meanwhile, LLoVi~\cite{Zhang2023ASL} unveiled an efficient, LLM-centric framework tailored for addressing queries that span long video durations. Here, video captioning agents transcribe videos into detailed textual descriptions which the LLMs then distill to enhance long-duration video comprehension.
While the aforementioned methodologies primarily translate video into text for LLM processing, LLMs are concurrently being explored for their capacity to facilitate video analysis through program generation.
ViperGPT~\cite{suris2023vipergpt} is a pioneering example, harnessing code-producing LLMs, including the likes of GPT-3 Codex~\cite{chen2021evaluating}. It effectively utilizes a visual module API catering to text-based queries and crafts programs that scrutinize image or video content, furnishing informed responses to those queries.
Similarly, ProViQ~\cite{Choudhury2023ZeroShotVQ} engages an LLM to craft Python scripts that enact multi-stage procedural reasoning in the context of zero-shot video queries, processing these scripts to ascertain solutions to posed questions.

\section{Method \& Analysis}
Adapting image MLLMs into the video domain can be tricky and vulnerable to the designs of model structures. 
In this section, we first present some challenges encountered when extending image MLLMs to video, drawing insights from our comprehensive experiments and analyses. Corresponding solutions to these challenges will be presented, forming the integral framework of \nameofmethod.

\subsection{Failure Cases Analysis for Applying Image MLLMs}
\label{subsec:failure_cases}
We first explored a direct way to adapt image MLLMs into the video domain: encoding selected video frames with image encoders separately and concatenating these frame features as input to the image MLLMs. 
This is to utilize the capability of LLMs to interpret the temporal information in the encoded video frames. 
%
We coined this method as \textit{n-frame}. Specifically, given a sequence of video frames $\mathbf{X}\in\mathbb{R}^{T\times C\times W \times H}$, we obtain the features for each frame via the vision encoder pre-trained in CLIP-ViT~\cite{radford2021learning} models and the encoded frames features are represented as $X_v\in \mathbb{R}^{T\times w \times h \times d}$. The \textit{n-frame} method is formulated as:
\begin{equation}
r = \text{MLLM}(X_v,X_t),
\end{equation}
where $X_t$ is the text inputs and r is the output texts. Nonetheless, during our efforts to train the MLLM in this scenario, we encountered two issues that hindered us from achieving optimally performance models.
%

\paragraph{Vulnerability to prompts.}

\begin{figure}[t]
    \centering
    \includegraphics[width=0.98\linewidth]{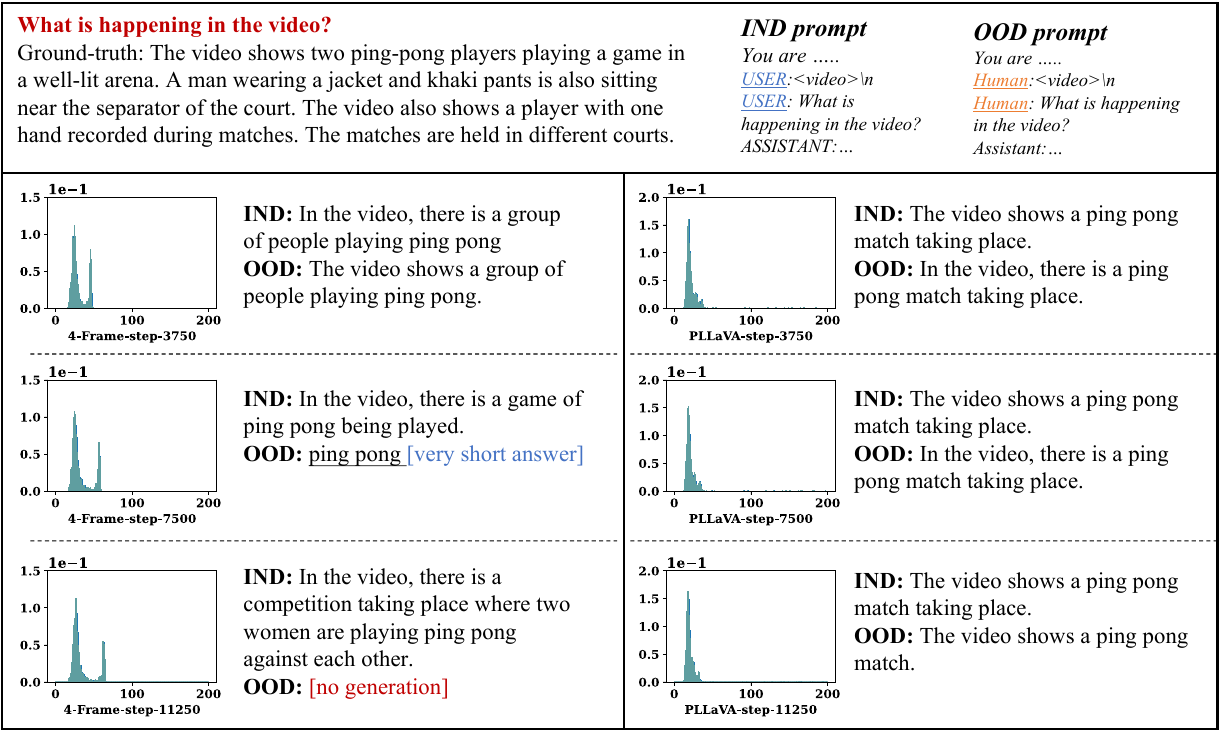}
    \caption{A example comparing the norm distributions and generated texts of the \textit{n-frame} and our \nameofmethod. From top to down, dominant tokens(with high norms) appear and increase as more data sample trained under the \textit{n-frame} setting, which is accompanied by a decline in generation quality, especially under the out-of-distribution prompt. On the right column, our \nameofmethod's norm distributions are consistent as well as the generated texts under various number of training data and prompts. 
    } 
    \label{fig:norm_hist_w_gen}
\end{figure}

The first observation is that the models trained with \textit{n-frame} could be highly sensitive to prompt patterns when dealing with generation tasks. \autoref{fig:norm_hist_w_gen} illustrates such a phenomenon. We divide the prompts into two categories: in-distribution (IND) and Out-of-Distribution (OOD). In the left part of the figure, when generating under the prompt pattern used in training (IND), the model can generate decent descriptions about the video despite its tendency of shorter generation length with more data samples trained. However, if we prompted the model with OOD prompts, in which we just changed the tags for the two roles in a conversation, the quality of the generated response then drastically declined. The generation has content in normal length under the model trained for 3750 steps. However, for the longer trained models, the generations are shorter for 7500 steps, and even no response for 11250 steps. This example demonstrate the vulnerability of the \textit{n-frame} method.

\paragraph{Dominant tokens.}

In view of the vulnerability of \textit{n-frame} models stated above, we proceeded to analyze the variance between models at their initial stages of training and when fully trained. By visualizing the norm of vision tokens across models trained at different stages, we observed a trend towards the emergence of dominant tokens( with high norms) as training samples increased, as shown by the histograms in~\autoref{fig:norm_hist_w_gen}. Furthermore, the twin-tower distribution is much wider when trained with more data. Therefore, we speculate there exists a plausible correlation between these dominant tokens and the degradation of generation under OOD prompt. The distribution comparisons between \textit{n-frame} and the proposed \nameofmethod can further validate the conjecture, which is explained in Sec.~\ref{sec:analysis}.

\paragraph{Data scaling failures.}
\begin{figure}
\centering
\subfigure[IND prompt.]{
    \begin{minipage}[b]{0.45\textwidth}
	\includegraphics[width=1\textwidth]{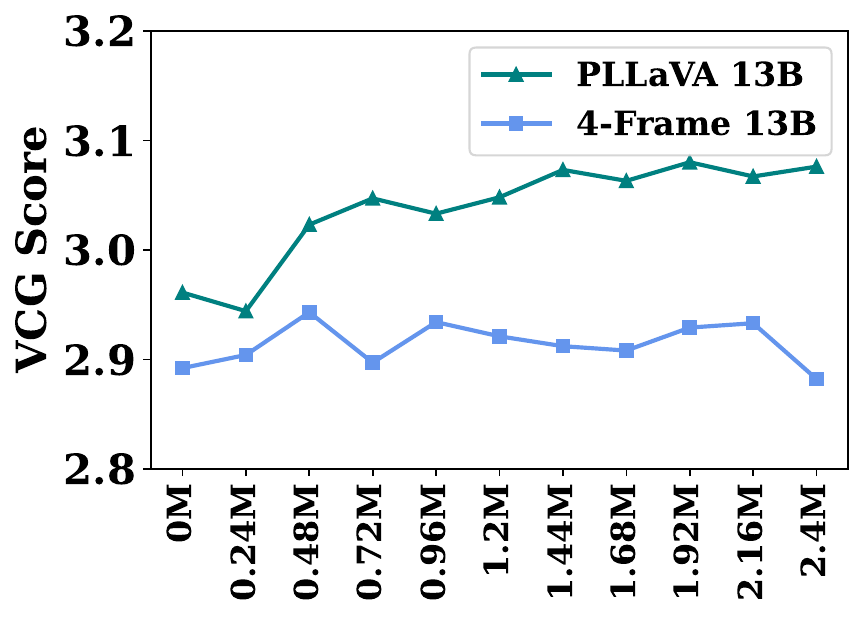}
    \end{minipage}
    \label{fig:data_scaling}
 }
\subfigure[OOD prompt.]{
\begin{minipage}[b]{0.45\textwidth}
    \includegraphics[width=1\textwidth]{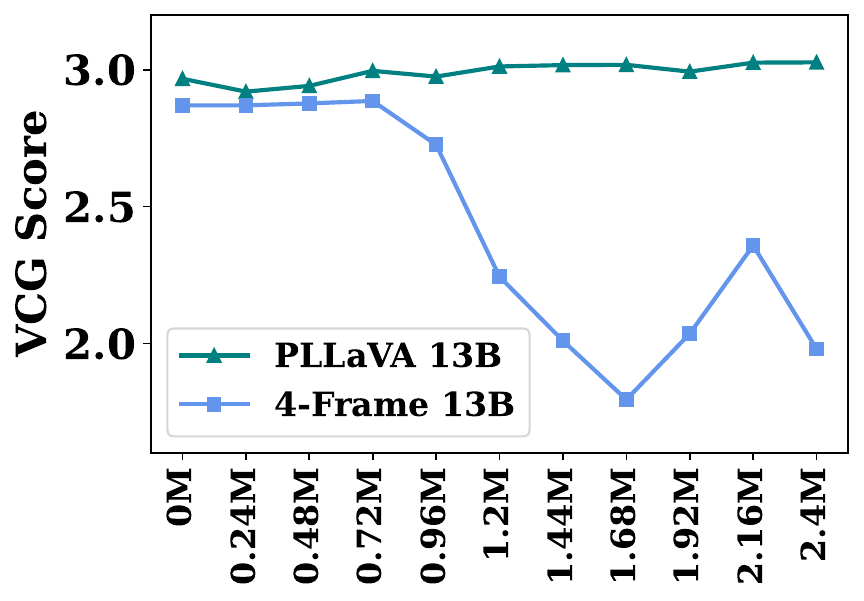}
    \end{minipage}
	\label{fig:data_scaling_ood}
    }
\caption{Validation curve of baseline model and PLLaVA. When training with more samples, (a) shows the validation curve with in-distribution (IND) prompt and (b) show the validation curve with out-of-disribution (OOD) prompts. It is observed that \textit{n-frame} method saturates fast and the performance even drops when training with longer time. is stagnant while pooling can keep rising. Differently, the performance of \nameofmethod{} keeps increasing during the training and saturates at the end of training.}
\label{fig:data_scaling_curve}
\vspace{-10pt}
\end{figure}

\begin{wraptable}{r}{0.5\textwidth}
\vspace{-10pt}
\scalebox{0.9}{
\begin{tabular}{lccc}
\toprule
\multirow{2}{*}{Method} & \multicolumn{3}{c}{Video-ChatGPT} \\
\cmidrule{2-4}
                        & reported  & reproduce  & scaled \\
\midrule           
Dataset                 & 100K  & 100K  & 100K+249K \\
            VCG Score   & 2.38      & 2.41       &  1.94      \\
\bottomrule
\end{tabular}
}
\caption{\label{tab:videochatgpt_exp} Video-ChatGPT~\cite{maaz2023videochatgpt} fails in data scaling.}
\vspace{-10pt}
\end{wraptable}

Based on the aforementioned phenomena, it can be inferred that employing image MMLMs to video and seeking to benefit from the scaling of video data samples can pose a challenging issue. We present the performance curve of \textit{n-frame} method under different training samples in~\autoref{fig:data_scaling_curve}. The blue curve representing performance tendency of \textit{n-frame} keeps stagnant under IND prompt, and degrades a lot under OOD prompts after the training sample exceeds 0.48M. Similar patterns are observed in the experimental findings of Video-ChatGPT~\cite{maaz2023videochatgpt}, as detailed in~\autoref{tab:videochatgpt_exp}. Video-ChatGPT~\cite{maaz2023videochatgpt} introduces a unique pooling strategy that involves averaging visual features across the temporal dimension as well as the spatial dimension, resulting a visual feature $X_{vcg}\in \mathbb{R}^{(T+w\times h)\times d}$ after concatenating both dimensions. This feature is then fed into LLMs to generate corresponding response. The first two columns of~\autoref{tab:videochatgpt_exp} demonstrate our replication of Video-ChatGPT using their 100K video-text dataset, while the third column illustrates a significant deterioration in model performance upon introducing additional training video data samples from VideoChat2~\cite{li2023mvbench}. 
Consequently, the identification of effective strategies for models to harness the growing volume of data remains a critical issue.

\subsection{Model Scaling Degradation}
\begin{wrapfigure}{r}{0.42\textwidth}
\vspace{-15pt}
\centering
\includegraphics[width=0.32\textwidth]{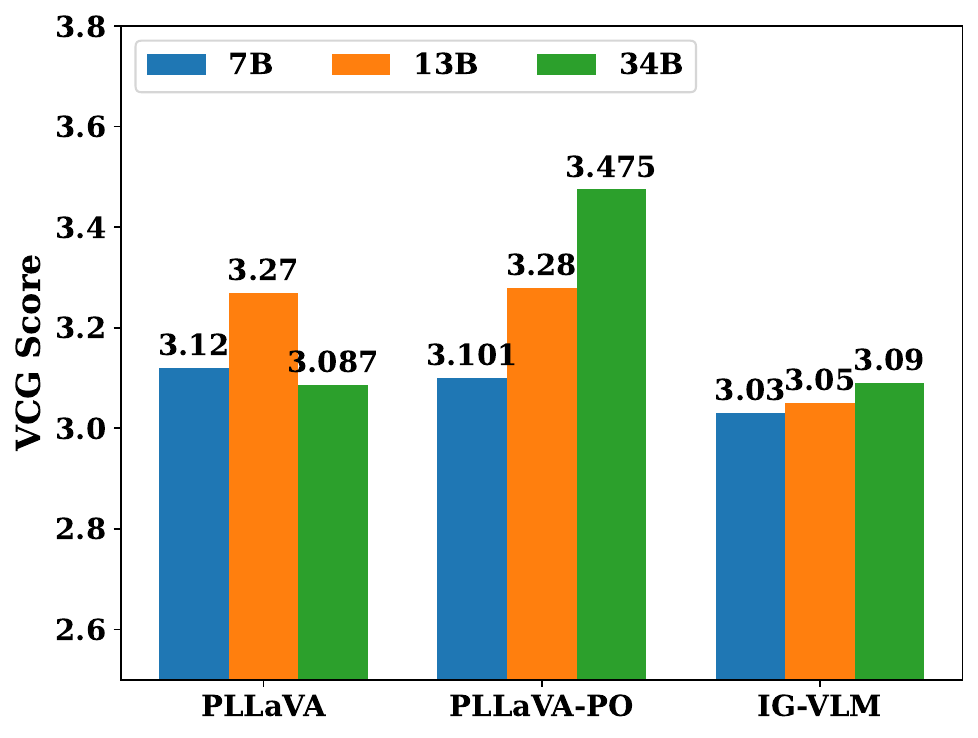}
\vspace{-10pt}
    \caption{Video MLLMs fail to improve when scaling model size. Post Optimization resolves the scaling degradation. 
    }
    \label{fig:scaling_curves}
    \vspace{-20pt}
\end{wrapfigure}
\label{phe:model_scaling_fail}
Our investigation on current video models reveals that increasing the model size does not typically result in significant improvements in performance for most models. We draw the performance of a recent work IG-VLM~\cite{kim2024igvlm} and our attempts in~\autoref{fig:scaling_curves}. IG-VLM achieves almost no difference when applying 7B, 13B, and 34B models of LLaVA-Next~\cite{liu2024llavanext}. In our attempts of with pooling features (the first column of~\autoref{fig:scaling_curves}), the performance of LLaVA-Next 34B is even worse than its 13B LLaVA-Next model. For IG-VLM, the input video frames are combined to a grid view image, confined by the resolution, leading to the unsatisfactory scaling ability. As for our attempts, we found a tendency of shorter generations with larger MLLMs, thus we owe the degradation to the quality of video-text data pairs, which undermines the generation ability of LLMs in MLLM models. 



\subsection{\nameofmethod}

\begin{figure}[!th]
    \centering
    \includegraphics[width=0.98\linewidth]{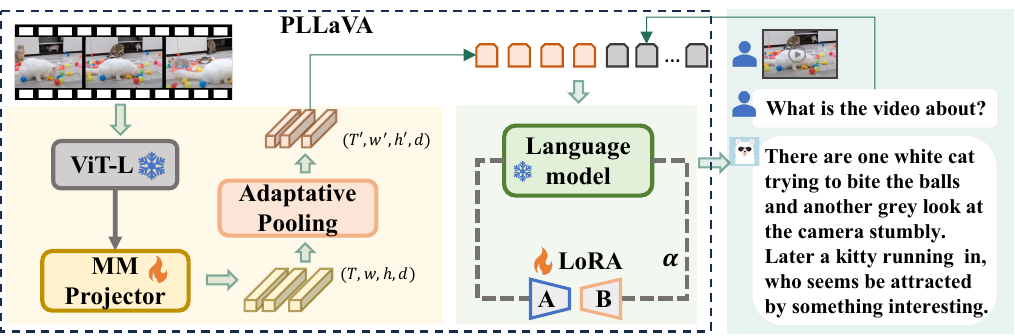}
    \caption{
    The framework of \nameofmethod begins with processing a video from the user through ViT-L and MM projector, yielding visual features with shape $(T, w, h, d)$. These features undergo average pooling, which effectively reduces both temporal and spatial dimensions. The pooled features are then flattened and concatenated with question embeddings, serving as input to the image Large Language Model to generate response to the user. The weights of the image LLMs are fused with LoRA weight learned under video samples. 
    } 
    \label{fig:framework}
\end{figure}

\paragraph{Motivation}
Our initial attempts on \textit{n-frame} and VideoChatGPT~\cite{maaz2023videochatgpt} reveal the intricacies of adapting image-focused MLLMs to the video domain, encountering the data scaling problem.  The former introduces a small amount of frames due to the limit of memory, whereas the latter compresses over 100 frames of information with pooling strategy. However, similar outcomes occur to both situations. 

In view of the necessity of temporal information and the prohibited costs of dealing with very long video input to MLLMs, pooling is an intuitive and simple way to fulfill both of the requirements.  The above two problems may stem from inadequacy of frame information and mishandling on the frame features. Therefore, in this paper, we deeply look into the pooling strategies for video features used in MLLMs.
\paragraph{Definition}

We formalize the pooling process for video features as follows, a model structure is shown in ~\autoref{fig:framework}. After feeding video frames $\mathbf{X}\in\mathbb{R}^{T\times C\times W \times H}$ into the CLIP-ViT model and the multimodal projector, we obtain an encoded vision feature $X_v\in \mathbb{R}^{T\times w \times h \times d}$ for a video input. This feature is then passed through a parameter-free Adaptive Average Structure Pooling module and reduced to a smaller size. Given the desired feature dimension $T'\times w'\times h'$,  the process is formulated as:

\begin{equation}
    X_{vp} = \text{AdaptStructPooling}(X_v|T'\times w'\times h').
\end{equation}

These features are then concatenated with text input embeddings and fed into the LLM to generate responses. We also include a LoRA~\cite{hu2021lora} module to adapt the LLM to video-related generation tasks. In conclusion, the trainable weights include Multimodal Projector and LLM LoRA. 

Within this framework, we investigated the impact of pooling through grid search analysis. Our findings suggest that pooling on the spatial dimension yields favorable outcomes, whereas temporal dimension pooling is associated with decreased performance. For a thorough exploration of our search process and the rationale behind this conclusion, please refer to Sec.~\ref{sec:pooling_influence}.

\subsection{Post Optimization}

Regarding the problem of performance decline associated with scaled model size, such degradation may stem from diminished language proficiency resulting from training on low-quality video-text data samples. To mitigate this, we propose a post-training optimization approach for the parameters of the video MLLM. It involves blending the trained Language Model (LLM) on video data with the original LLM of the base image MLLM. For a pretrained MLLM with LLM parameters  $W_0$ and given input $X_{vp}$, the output hidden states from the LoRA fine-tuned LLM can be acquired as follows:
\begin{equation}
h = W_0X_{vp}+\frac{\alpha}{r}\Delta WX_{vp},
\end{equation}
where $\Delta W$ are a low-rank learnable parameters for adapting $W_0$, and $\frac{\alpha}{r}$ is used to scale the learned low-rank weight.

As part of our post-training optimization process, we tune the mix ratio between the original LLMs and the trained LLMs (incorporating LoRA weights) by varying the value of $\alpha$ during inference. Our experiments indicate that lower $\alpha$ yields significantly better generative performance.




\section{Experiments}

\subsection{Experiment Setting}
\paragraph{Data and Evaluation}
We leverage instructional video-to-text datasets to extend the capabilities of image MLLMs to handle video inputs. The training data are sourced from the dataset used in VideoChat2~\cite{li2023mvbench}, which embraces data for various video understanding tasks, including 27k conversation videos from VideoChat~\cite{li2023videochat} and Video-ChatGPT~\cite{maaz2023videochatgpt}, 80k data of classification tasks from Kinetics~\cite{kay2017kinetics} and SthSthV2~\cite{goyal2017sthsthv2}, 450k captioned data from Webvid~\cite{bain2021webvid}, YouCook2~\cite{zhou2018youcook2}, TextVR~\cite{wu2023textvr} and VideoChat, 117 reasoning data from NextQA~\cite{xiao2021nextqa} and CLEVRER~\cite{yi2020clevrer} and 109K annotated questioning answering data samples from Webvid, TGIF~\cite{li2016tgif} and Ego4D~\cite{Grauman2021Ego4DAT}. In total, we use 783k instructional tuning data. 

We evaluate our trained models with the following video-to-text benchmarks. First, the open-ended Video Question Answer (VideoQA) includes MSVD-QA~\cite{xu2017videoqa}, MSRVTT-QA~\cite{xu2017videoqa}, ActivityQA~\cite{yu2019activitynet}, and TGIF QA~\cite{li2016tgif}. Responses in these question-answering benchmarks typically consist of single-word answers. GPT-3.5 \cite{chatgpt} is used to evaluate the accuracy (Accuracy, with answers true/false) and quality (Score, ranging from 0 to 5) of the models' responses.
Additionally, we adopt the Video-based Generative Performance benchmark (referred to as VCG Score), introduced by VideoChatGPT~\cite{maaz2023videochatgpt}. These benchmarks often involve longer answers, encompassing five aspects of video understanding: CI (Correctness of Information), DO (Detail Orientation), CU (Context Understanding), TU (Temporal Understanding), and CO (Consistency). The generation is also assessed using the GPT-3.5 model.
Furthermore, we also use the multi-choice Question Answering benchmark, MVBench~\cite{li2023mvbench}, comprising 20 tasks that demand nuanced temporal comprehension of videos. This benchmark does not necessitate evaluation from the GPT-3.5 model.

\paragraph{Models and Implementation Details}
\nameofmethod is constructed upon the image MLLMs, LLaVA Next~\cite{liu2024visual,liu2024llavanext} models 7B, 13B, and 34B. We utilize their pre-trained weights available in the Hugging Face library\footnote{https://huggingface.co/docs/transformers/en/model\_doc/llava\_next} and integrate an average pooling to reduce feature dimensions before passing the input visual features to the LLM generation component.
For the pooling layer, we uniformly sample 16 frames as input and set the target pooling shape to be $16\times12\times12\times d$, where $d$ corresponds to the input dimension of the LLMs. 
During training, we employ a batch size of 128 and a learning rate of 2e-5, with a cosine scheduler and a warmup ratio of 0.03. All the reported results are evaluated on models trained for 6250 steps.
For evaluation, we adopt the GPT-3.5-turbo-0125 model across all the benchmarks. 

\subsection{Impact of Pooling Operation Design}
\label{sec:pooling_influence}
Considering the unsatisfying performance of the complete pooling on temporal and spatial dimensions adopted in Video-ChatGPT and the limitation information in the straightforward \textit{n-frame} method, we further explore the influence of poling strategies here. 
\paragraph{Pooling Layer Design}
\label{para:pooling_hyper}
\begin{figure}[ht]
\centering

\subfigure[Spatial shape effects on MVBench.]{
    \begin{minipage}[b]{0.45\textwidth}
        \includegraphics[width=1\textwidth]{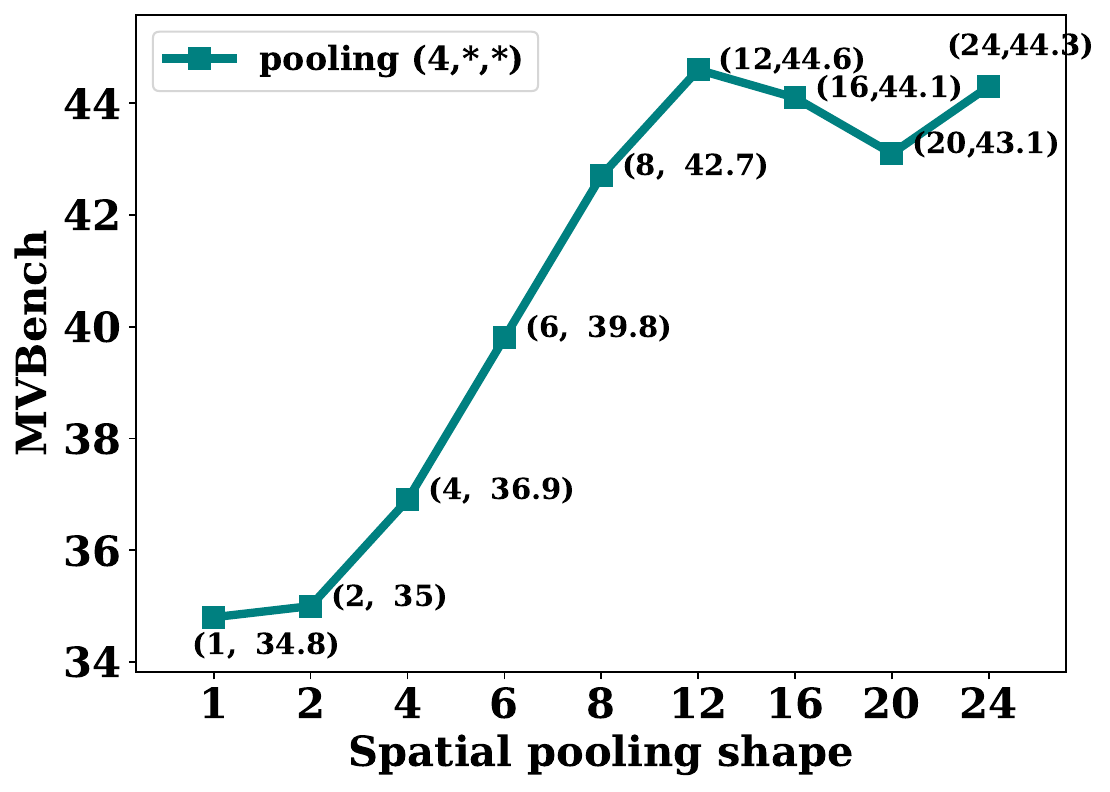}
    \end{minipage}
    \label{subfig:spatial_effect_mvbench}
    }
\subfigure[Spatial shape effects on VCG.]{
    \begin{minipage}[b]{0.45\textwidth}
        \includegraphics[width=1\textwidth]{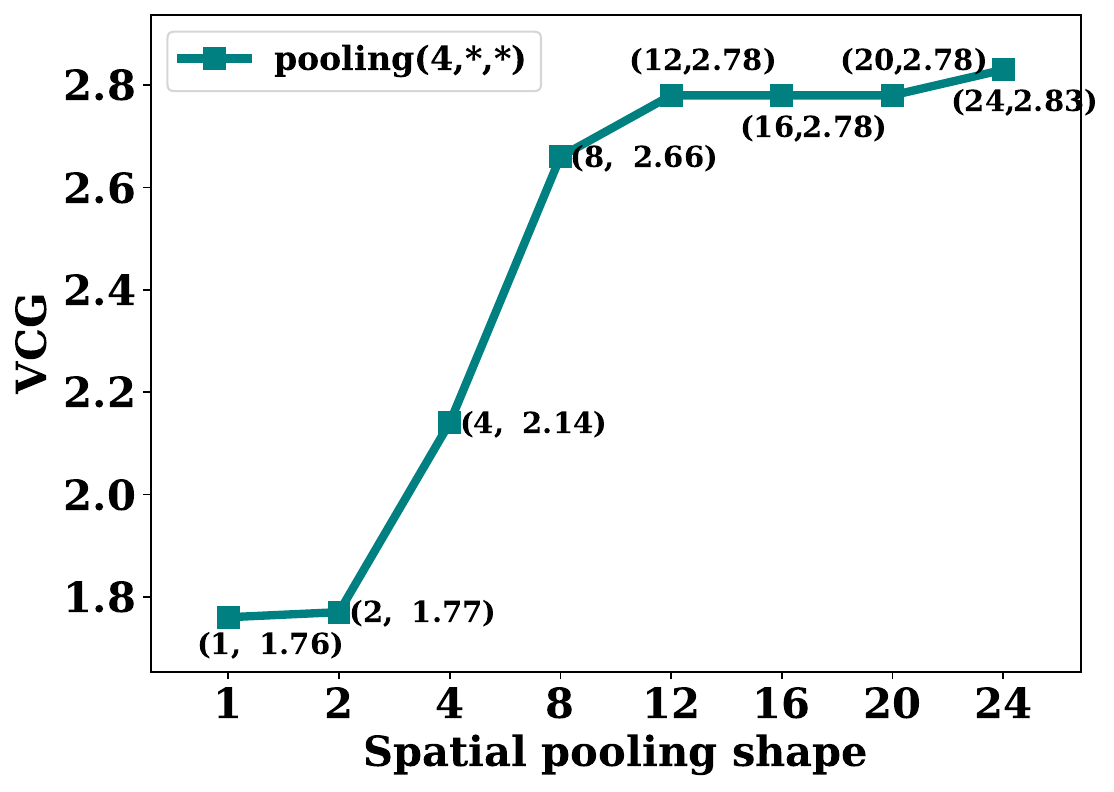}
    \end{minipage}
    \label{subfig:spatial_effect_vcg}
    }
\subfigure[Temporal shape effects on MVBench.]{
    \begin{minipage}[b]{0.45\textwidth}
        \includegraphics[width=1\textwidth]{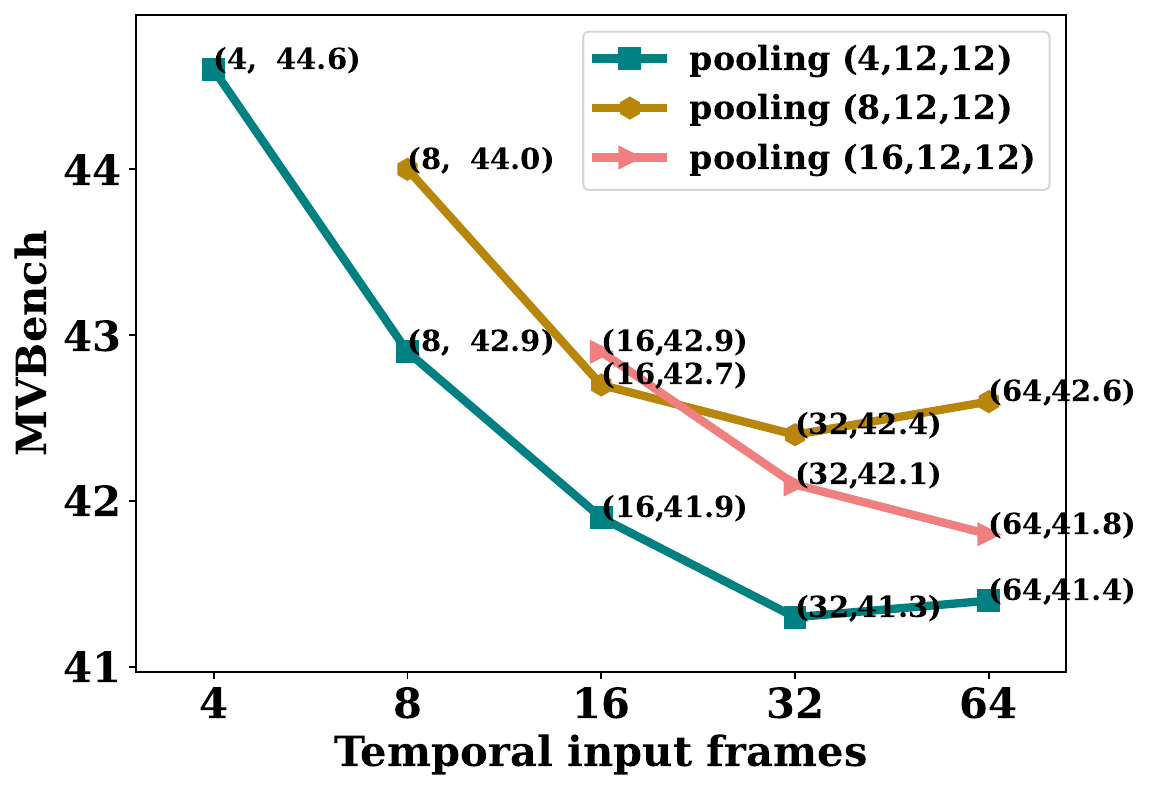}
    \end{minipage}
    \label{subfig:temporal_effect_mvbench}
    }
\subfigure[Temporal shape effects on VCG.]{
    \begin{minipage}[b]{0.45\textwidth}
        \includegraphics[width=1\textwidth]{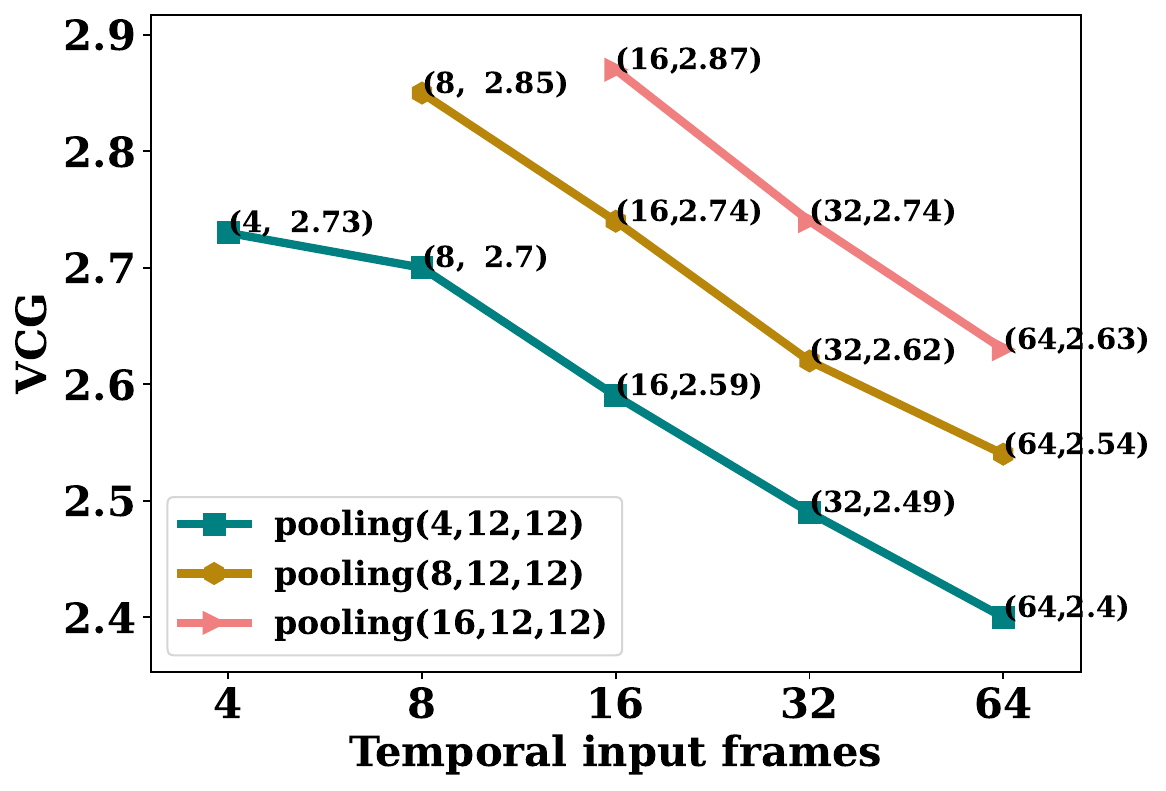}
    \end{minipage}
    \label{subfig:temporal_effect_vcg}
    }
\caption{Pooling shape influence.}
\label{fig:pooling_shape}
\end{figure}

Pooling can be done both temporally and spatially. In this part, we aim to figure out the answer to two questions: 1) which dimension is more suitable to be pooled to save the computational cost and 2) what is the largest compression ratio along that dimension. To achieve this, we plot a model curve based on the LLaVA-1.5 7B model with different temporal and spatial dimensions controlled via pooling operation. 
%
%
Specifically, for the spatial dimension, we picked an input video feature with shape (4,24,24,$d$), where 4 is the frame numbers (temporal dimension), 24$\times$24 is the original spatial dimension of frame features, and $d$ is the embedding dimension of each visual token. The target spatial shapes are chosen at evenly spaced intervals between 1 and 24, resulting in a set of spatial shapes $S=$\{$n\times n$ | $n \in [1,2,4,6,8,12,16,20,24$]\}. The MVBench and VCG Score performance of these spatial pooling shapes are shown in Figure~\ref{subfig:spatial_effect_mvbench} and~\ref{subfig:spatial_effect_vcg}. It is observed that downsampling the spatial dimension by 50\% does not degrade the model performance. Further reducing the spatial dimension would lead to a significant performance drop. Considering the tradeoff between computational overhead and performance, 12$\times$12 can be a target spatial dimension.

We further experimented on the temporal dimension. Several target pooling shapes were chosen with spatial dimensions fixed as 12, including (4,12,12), (8,12,12), and (16,12,12). We study the pooling performance tendency when altering the number of input video frames, indicating the downsampling rate of poolings. For example, pooling from (64,24,24) to (4,12,12) indicates every 16 frames are fused, then the downsampling rate should be 6.25\%. All of the resulting model curves are shown in Figure~\ref{subfig:temporal_effect_mvbench} and ~\ref{subfig:temporal_effect_vcg}. Different from spatial pooling, the model performance is sensitive to the temporal pooling. As illustrated in these two figures, all lines achieve better performance with lower downsmapling rates.
In other words, \emph{pooling along temporal dimension always downgrades the model performance.}

\paragraph{Pooling Impact}
We found that pooling over more video frames not only improves the model efficiency but also makes the model more robust to user enquires. 
During our experiments, we evaluated models under different training iterations with two sets of prompts. For example, we vary the role tag from `USER' to `Human' during evaluation and the results are as shown in~\autoref{fig:norm_hist_w_gen}. The figure shows that the visual feature norms learned with the pooling operation show consistent distributions under different training iterations compared to the 4-frame method that shows dominant tokens. 
This is also reflected in the model responses where the pooling method gives consistent good text responses while the 4-frames method gives shorter and shorter text responses as the training goes longer, or even no response when out-of-distribution prompts are used.
%
This conclusion can be further validated by~\autoref{fig:text_length_compare}. With pooling introduced, no matter what prompt is used or how much training sampled is learned, the text generation lengths with the pooling method are consistent. We owe the stability in generation to the smoothing ability of pooling, which eliminates the influence of dominant high norm tokens. For more rigorous analysis from the perspective of mathematical proofs, we leave it for future work. 

\subsection{Quatitative Results}

\begin{table}[h]
\centering
\scalebox{0.58}{
\begin{tabular}{l ll ll ll ll ll llllll}

\toprule
\multirow{2}{*}{\textbf{Method}} & \multirow{2}{*}{\makecell{\textbf{Vision}\\\textbf{Encoder}}} & \multirow{2}{*}{\makecell{\textbf{LLM}\\\textbf{Size}}} & \multicolumn{2}{c}{\textbf{MSVD-QA}} & \multicolumn{2}{c}{\textbf{MSRVTT-QA}} & \multicolumn{2}{c}{\textbf{ActivityNet-QA}} & \multicolumn{2}{c}{\textbf{TGIF-QA}} & \multicolumn{6}{c}{\textbf{Video-ChatGPT}} 
\\ 
\cmidrule(lr){4-5} \cmidrule(lr){6-7}  \cmidrule(lr){8-9} \cmidrule(lr){10-11} \cmidrule(lr){12-17}  
& & &\textbf{~Acc.~} & \textbf{Sco.} & \textbf{~Acc.~} & \textbf{Sco.} & \textbf{~~Acc.~~} & \textbf{Sco.} & \textbf{~Acc.~} & \textbf{Sco.} & \textbf{CI} & \textbf{DO} & \textbf{CU} & \textbf{TU} & \textbf{CO} & \textbf{Avg.}\\
\midrule
FrozenBiLM\cite{Yang2022FrozenBiLM}& ViT-L & 1.3B & 33.8 & - & 16.7 & - & 25.9 & - & 41.9 & - \\
Video-LLaMA\cite{zhang2023VideoLLAMA} & CLIP-G & 7B    & 51.6 & 2.5 & 29.6 & 1.8 & 12.4 & 1.1 & - & - &  1.96 & 2.18  & 2.16 & 1.82  & 1.79  & 1.98\\
LLaMA-Adapter\cite{Zhang2023LLaMAAdapterEF} & ViT-B & 7B    & 54.9 & 3.1 & 43.8 & 2.7 & 34.2 & 2.7 & - & - & 2.03 & 2.32 & 2.30 & 1.98 & 2.15 & 2.16\\
Video-ChatGPT\cite{maaz2023videochatgpt} & ViT-L & 7B   & 64.9 & 3.3 & 49.3 & 2.8 & 35.2 & 2.7 & 51.4 & 3.0 & 2.50 & 2.57 & 2.69 & 2.16 & 2.20  & 2.42\\
Video-LLaVA\cite{Lin2023VideoLLaVALU} & ViT-L & 7B   & 70.7 & 3.9 & 59.2 & 3.5 & 45.3 & 3.3 & 70.0 & 4.0 \\
Chat-UniVi\cite{Jin2023ChatUniViUV}   & ViT-L & 7B   & 65.0 & 3.6 & 54.6 & 3.1 & 45.8 & 3.2 & 60.3 & 3.4 & 2.89 & 2.91 & 3.46 & 2.89 & 2.81 & 2.99\\
MovieChat\cite{Song2023MovieChatFD}   & CLIP-G  & 7B   & 75.2 & 3.8 & 52.7 & 2.6 & 45.7 & 3.4 & - & - & 2.76  & 2.93  & 3.01 & 2.24 & 2.42 & 2.67\\
VideoChat\cite{li2023videochat}   & CLIP-G  & 7B   & 56.3 & 2.8 & 45.0 & 2.5 & 26.5 & 2.2 & 34.4 & 2.3 & 2.23  & 2.50  & 2.53 & 1.94 & 2.24 & 2.29\\
VideoChat2\cite{li2023mvbench}  & UMT-L & 7B   & 70.0   & 3.9 & 54.1 & 3.3 & 49.1 & 3.3 & - & - & 3.02  & 2.88 & 3.51 & 2.66 & 2.81 & 2.98 \\
Vista-LLaMA\cite{Ma2023VistaLLaMARV} & CLIP-G  & 7B   & 65.3 & 3.6 &  60.5 & 3.3 & 48.3 & 3.3 & - & - & 2.44  & 2.64 & 3.18 & 2.26 & 2.31 & 2.57  \\ 
LLaMA-VID\cite{Li2023LLaMAVIDAI}   & CLIP-G  & 13B   & 70.0   & 3.7 & 58.9 & 3.3 & 47.5 & 3.3 & - & - & 2.96  & 3.00  & 3.53 & 2.46 & 2.51 & 2.89 \\

LITA~\cite{huang2024lita} & CLIP-L & 7B & - & - & - & - & - & - & - & - & 2.94 & 2.98 & 3.43 &  2.68 & 3.19 & 3.04 \\

ST-LLM~\cite{liu2024st} & BLIP2 & 7B & 74.6  & 3.9 & 63.2 & 3.4 & 50.9 & 3.3 & - & - & 3.23 & 3.05 & 3.74 &  2.93 & 2.81 & 3.15 \\

IG-VLM CogAgent\cite{hong2023cogagent} & CLIP-E & 7B    & 76.7 & 4.1 & 62.7 & 3.6 & 57.3 & 3.6 & 76.7 & 4.0 & 3.26 & 2.76 & 3.57 & 2.34 & 3.28 & 3.04\\

IG-VLM LLaVA 7B \cite{liu2024llavanext}  & ViT-L  & 7B  & 78.8 & 4.1 & 63.7 & 3.5 & 54.3 & 3.4 & 73.0 & 4.0 & 3.11 & 2.78 & 3.51 & 2.44 & 3.29 & 3.03\\
IG-VLM LLaVA 13B \cite{liu2024llavanext} & ViT-L & 13B  & 77.4 & 4.1 & 62.6 & 3.4 & 57.1 & 3.5 & 78.0 & 4.0 & 3.17 & 2.79 & 3.52 & 2.51 & 3.25 & 3.05 \\
IG-VLM LLaVA 34B \cite{liu2024llavanext} & ViT-L & 34B  & 79.6 & 4.1 & 62.4 & 3.5 & 58.4 & 3.5 & 79.1 & 4.2 & 3.21 & 2.87 & 3.54 & 2.51 & \textbf{3.34} & 3.09\\
\midrule

IG-VLM GPT-4V\cite{achiam2023gpt} & Unk & GPT-4  & 76.3 & 4.0 &  63.8 & 3.5 & 57.0 & 3.5 & 65.3 & 3.7 & 3.40 & 2.80 & 3.61 & \textbf{2.89} & 3.13 & {3.17} \\
\midrule
\nameofmethod 7B  & ViT-L & 7B   & 76.6 & 4.1 & 62.0 & 3.5 & 56.3 & 3.5 &  77.5 & 4.1  & 3.21 & 2.86 & 3.62 & 2.33 & 2.93 & 3.12 \\
\nameofmethod 13B & ViT-L & 13B  & 75.7 & 4.1 & 63.2 & 3.6 & 56.3 & 3.6 &  77.8  & 4.2 & 3.27 & 2.99 & 3.66 & 2.47 & 3.09 & 3.27\\
\nameofmethod 34B & ViT-L & 34B  &  \textbf{79.9} & \textbf{4.2} & \textbf{68.7} & \textbf{3.8} & \textbf{60.9}& \textbf{3.7} & \textbf{80.6} & \textbf{4.3} & \textbf{3.60} & \textbf{3.20} & \textbf{3.90}& 2.67 & 3.25 & \textbf{3.48}\\
\midrule

Improve over GPT-4V \cite{kim2024igvlm} & - & - &  \greentext{+3.6} & \greentext{+0.2}  & \greentext{4.9} & \greentext{0.3} & \greentext{3.9} & \greentext{0.2} & \greentext{15.3} & \greentext{0.6} & \greentext{0.2} & \greentext{0.4} & \greentext{0.3} & \redtext{-0.32} & \greentext{0.12} & \greentext{0.31}\\
\bottomrule
\end{tabular}
}
\caption{Results of video question-answering. }
\label{tab:generative_benchmark}
\end{table}

\begin{table}[t]
\centering
\scalebox{0.48}{
\begin{tabular}{l cc ccccc ccccc ccccc cccccc }
\toprule
\textbf{Method} & \makecell{\textbf{Vision}\\\textbf{Encoder}} & \makecell{\textbf{LLM}\\\textbf{Size}} 
& \textbf{AS} & \textbf{AP} & \textbf{AA} & \textbf{FA} & \textbf{UA} 
& \textbf{OE} & \textbf{OI} & \textbf{OS} & \textbf{MD} & \textbf{AL} 
& \textbf{ST} & \textbf{AC} & \textbf{MC} & \textbf{MA} & \textbf{SC} 
& \textbf{FP}  & \textbf{CO} & \textbf{EN} & \textbf{ER} & \textbf{CI} & \textbf{Avg.}\\
\midrule
Video-LLaMA~\cite{zhang2023VideoLLAMA} & CLIP-G & 7B &  27.5 &25.5 & 51.0 & 29.0 & 39.0 & 48.0 & 40.5 & 38.0 & 22.5 & 22.5 & 43.0 & 34.0 & 22.5 & 32.5 & 45.5 & 32.5 & 40.0 & 30.0 & 21.0 & 37.0 & 34.1 \\
LLaMA-Adapter~\cite{Zhang2023LLaMAAdapterEF} & ViT-B & 7B & 23.0 & 28.0 & 51.0 & 30.0 & 33.0 & 53.5 & 32.5 & 33.5 & 25.5 & 21.5 & 30.5 & 29.0 & 22.5 & 41.5 & 39.5 & 25.0 & 31.5 & 22.5 & 28.0 & 32.0 & 31.7 \\
Video-ChatGPT~\cite{maaz2023videochatgpt} & ViT-L & 7B & 23.5 & 26.0&  62.0 & 22.5 & 26.5 & 54.0 & 28.0 & 40.0 & 23.0 & 20.0 & 31.0 & 30.5 & 25.5 & 39.5 & 48.5 & 29.0 & 33.0 & 29.5 & 26.0 & 35.5 & 32.7 \\

VideoChat~\cite{li2023videochat} & CLIP-G & 7B & 33.5 & 26.5 & 56.0 & 33.5 & 40.5 & 53.0 & 40.5 & 30.0 & 25.5 & 27.0 & 48.5 & 35.0 & 20.5 & 42.5 & 46.0 & 26.5 & 41.0 & 23.5 & 23.5 & 36.0 & 35.5 \\
VideoChat2~\cite{li2023mvbench} & UMT-L & 7B & 66.0 & 47.5 & \textbf{83.5} & \textbf{49.5} & 60.0 & 58.0 & \textbf{71.5} & \textbf{42.5} & 23.0 & 23.0 & 88.5 & 39.0 & 42.0 & 58.5 & 44.0 & 49.0 & 36.5 & 35.0 & 40.5 & \textbf{65.5} & 51.1 \\

ST-LLM~\cite{liu2024st} & BLIP2 & 7B  & 66.0 & 53.5 & 84.0 & 44.0 & 58.5 & 80.5 & 73.5 & 38.5 & 42.5 & 31.0 & 86.5 & 36.5 & 56.5 & 78.5 & 43.0 & 44.5 & 46.5 & 34.5 & 41.5 & 58.5 & 54.9 \\

\midrule

GPT-4V & Unk & GPT-4 & 55.5 & \textbf{63.5} & 72.0 & 46.5 & 73.5 & 18.5 & 59.0 & 29.5 & 12.0 & 40.5 & 83.5 & 39.0 & 12.0 & 22.5 & 45.0 & 47.5 & 52.0 & 31.0 & 59.0 & 11.0 & 43.5\\
\midrule
\nameofmethod 7B  & ViT-L & 7B  & 58.0 & 49.0 & 55.5 & 41.0 & 61.0 & 56.0 & 61.0 & 36.0 & 23.5 & 26.0 & 82.0 & 39.5 & 42.0 & 52.0 & 45.0 & 42.0 & 53.5 & 30.5 & 48.0 & 31.0 & 46.6 \\
\nameofmethod 13B & ViT-L & 13B & 66.0 & 53.0 & 65.5 & 45.0 & 65.0 & 58.0 & 64.5 & 35.5 & 23.5 & 30.0 & 85.0 & 39.5 & 45.5 & 57.0 & 47.5 & 49.5 & 49.0 & 33.0 & 53.0 & 37.0 & 50.1 \\
\nameofmethod 34B & ViT-L & 34B & \textbf{67.5} & 53.0 & 82.0 & 47.0 & \textbf{79.0} & \textbf{68.5} & 67.5 & 36.5 & \textbf{37.5} & \textbf{49.5} & \textbf{91.0} & \textbf{40.5} & \textbf{43.0} &\textbf{70.0} & \textbf{51.5} & \textbf{50.0} & \textbf{66.5} & \textbf{39.5} & \textbf{63.5} & 59.0 & \textbf{58.1} \\
\midrule
Improve over GPT-4V & - & -& \greentext{12.0} & \redtext{-10.5} & \greentext{10.0} & \greentext{1.5} & \greentext{5.5} & \greentext{50} & \greentext{8.5} & \greentext{7.0} & \greentext{25.5} & \greentext{9.0} & \greentext{7.5} & \greentext{1.5} & \greentext{31.0} &\greentext{57.5} & \greentext{5.5} & \greentext{2.5} & \greentext{14.5} &\greentext{8.5} & \greentext{4.5} & \greentext{48.0} & \greentext{14.5} \\
\bottomrule
\end{tabular}}
\caption{Results on MVBench multi-choice question answering.}
\label{tab:mvbench}
\end{table}

\autoref{tab:generative_benchmark} demonstrates the results on VideoQA. \nameofmethod 34B significantly outperforms all the existing methods on the Accuracy and Score metrics of MSVD, MSRVTT, ActivityNet, and TGIF. Compared to GPT-4V, \nameofmethod 34B achieves improvement margins of 3.6, 4.9, 3.9, and 15.3 on these four benchmarks. The performance of \nameofmethod with 7B and 13B model sizes also exceeds all the baselines on the Score metric. These results not only prove the capability of our model in conducting video question answering but also highlight the superiority of our pooling strategy in scaling model size.

\nameofmethod also achieved a new state-of-the-art in the average VCG score. The 7B, 13B, and 34B versions have all outperformed their best counterparts of the same LLM size, with margins of 2.9\%, 7.1\%, and 12.6\%, respectively. Notably, \nameofmethod achieves superior performance on CI(correctness of information), DO(Detail Orientation), and CU(Context Understanding) compared to the previous SOTA, with 34B exceeding them by 5.8\%, 6.7\%, 9.2\%. These results indicate that \nameofmethod will be of great potential to do detailed video captioning. As for TU(temporal understanding), \nameofmethod 34B exceeds its fair opponent IG-VLM LLaVA 34B by 6\%. Compared with models that utilize the specialized video encoder, VideoChat2, or a more complicated frame combination method, Chat-Univ, \nameofmethod still has some room for improvement by fingering the pooling strategy or incorporating a better vision encoder. CO(Consistency) measures generation consistency when the model encounters different questions that lead to similar answers. Compared to baselines except for IG-VLM, our model achieves much better consistency. 

MVBench is a comprehensive video understanding benchmark, focusing on questions that require overall comprehension of multiple frames. As shown in ~\autoref{tab:mvbench}, \nameofmethod surpasses the previous SOTA VideoChat2 with a margin of 13.7\% on average across 20 tasks. If we look into each aspect of MVBench, our method performs very well, concerning 17 out of 20 tasks of MVBench, which shows that our model has the superiority to understand many fine-grained details about videos accurately. However, we also noticed some aspects of our model still need to improve, such as CI(CounterFactual Inference) and OS(object shuffle). CI is used to predict what might happen if an event occurs, and OS is used to locate the final position of an object in an occlusion game. These two require strong reasoning ability and imagination to answer. VideoChat2 is pretrained with a large amount of video data with a specialized video encoder and fine-tuned with both video and image reasoning data, thus presenting better performance in these aspects.

\subsection{Analysis}
\label{sec:analysis}
Our \nameofmethod is a simple and parameter-efficient method to adapt image MLLMs into the video domain. We also provide a feasible way to scale the models to larger sizes, which we found is hard to achieve in other methods such as ChatUniv~\cite{Jin2023ChatUniViUV} and IG-VLM~\cite{kim2024igvlm}. In the following, we further provide some analysis related to the explanations on pooling shapes and the influence of LoRA weight on different tasks.

\paragraph{Temporal or spatial pooling?}
\begin{figure}[h]
    \centering
    \includegraphics[width=0.95\textwidth]{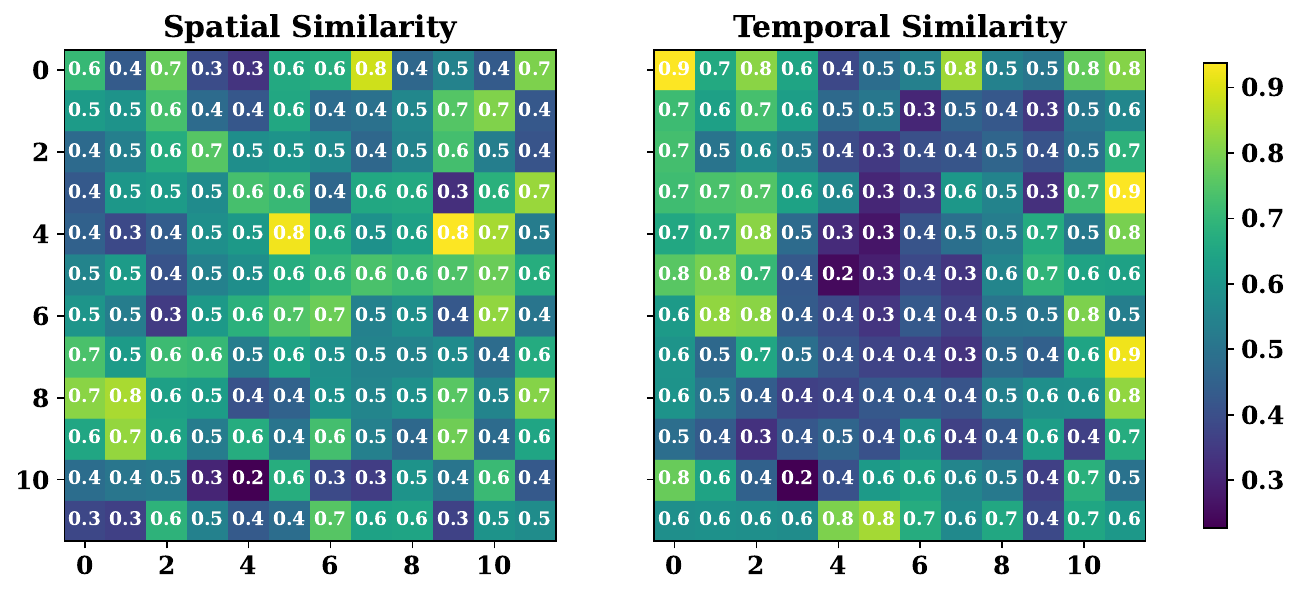}
    
    \caption{\label{fig:token_similarity}Vision token embedding similarities between spatial token neighbors and temporal token neighbors.}
\end{figure}
In Sec.\ref{para:pooling_hyper}, we have illustrated the impact of temporal and spatial poolings, concluding that pooling along the temporal dimension consistently results in decreased performance compared to retaining the original frame numbers. We attribute this phenomenon to the interference with token features. In image MLLMs, features are derived from images/video frames using CLiP-ViT models, which produce embedded patches for each image/video frame, resulting in a video feature with shape $(T, H, W)$. Pooling changes the dimensions of $T$ (time), $H$ (height), and $W$ (weight). In contrast to pooling along the spatial dimension (local pooling on single images/frames, changing $H$ and $W$), pooling along the temporal dimension (changing $T$) risks altering the original frame features. To validate the guess, we visualize token similarities among spatial and temporal token neighbors for a video feature in \autoref{fig:token_similarity}. The two subfigures reveal significantly higher similarities within spatial neighbors than temporal neighbors. This observation supports the potential distortion of original token features caused by temporal pooling. LLMs are designed for sequence understanding. Even without preprocessing on temporal information aggregation, they can model temporal relations.

\paragraph{Image? Video? or Both?}
\begin{figure}
    \centering
\subfigure[]{
    \begin{minipage}[b]{0.4\textwidth}
        \includegraphics[width=1\textwidth]{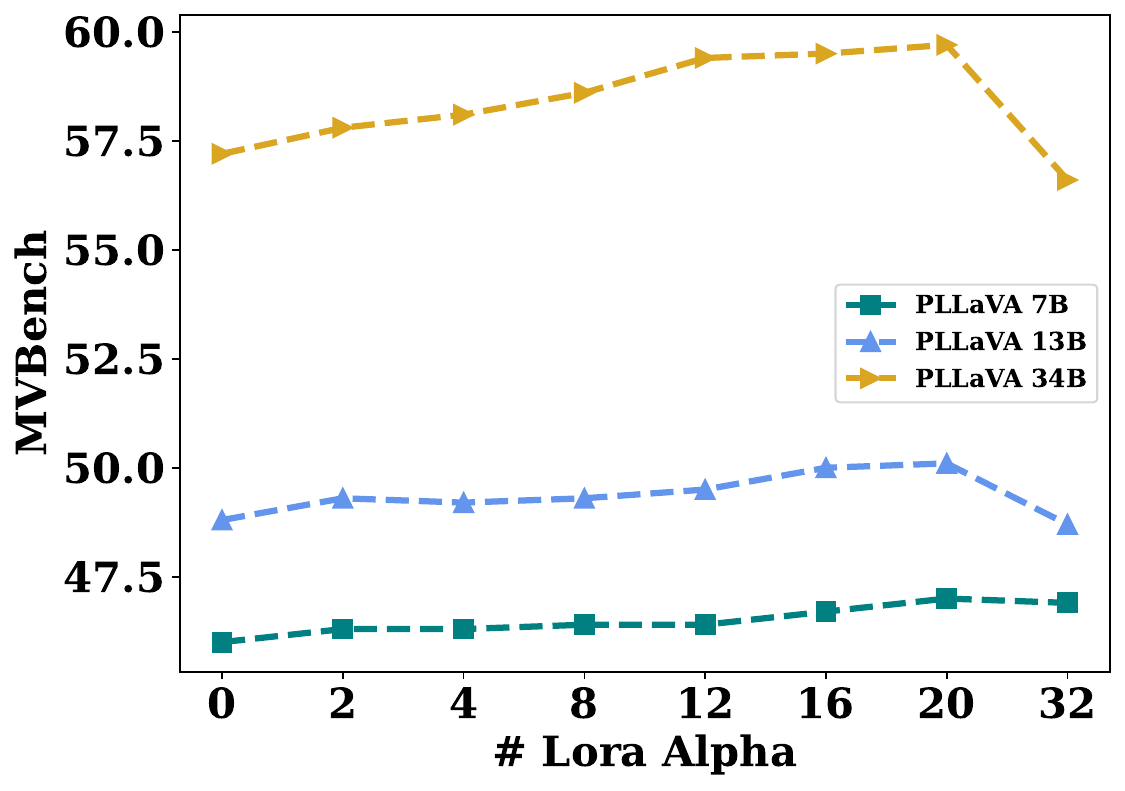}
    \end{minipage}
    \label{subfig:mvbench_with_alpha}
    }
\subfigure[]{
    \begin{minipage}[b]{0.4\textwidth}
        \includegraphics[width=1\textwidth]{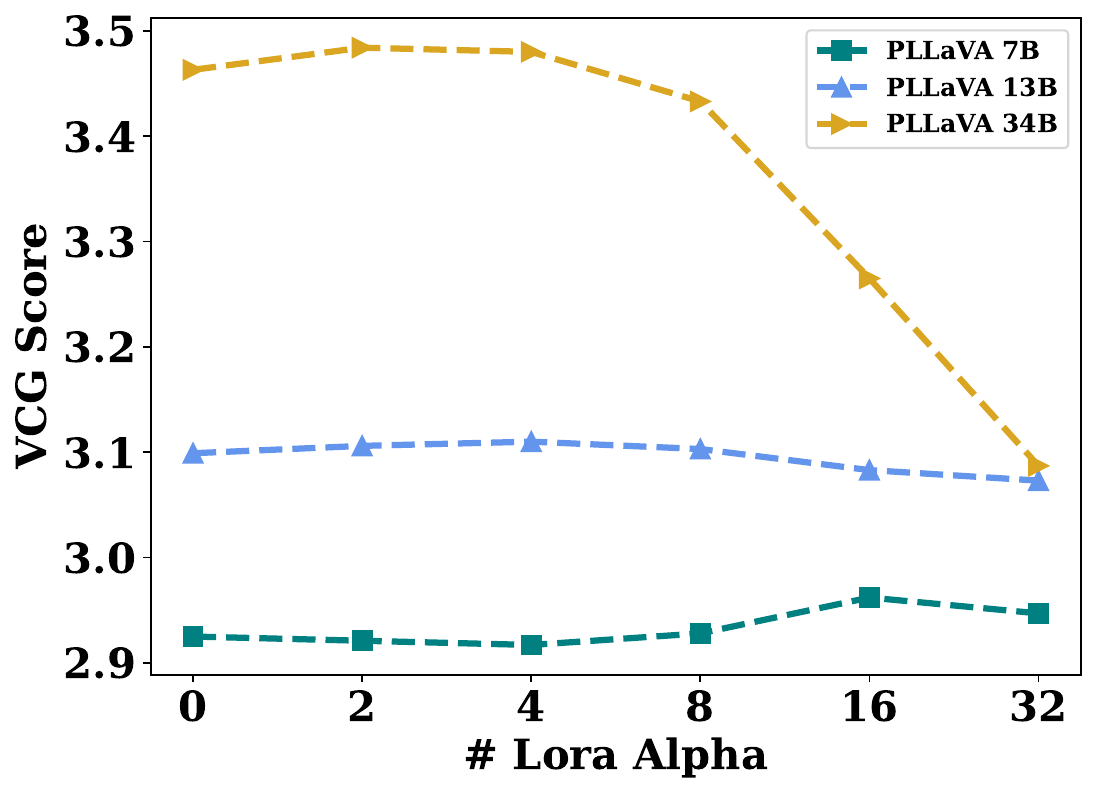}
    \end{minipage}
    \label{subfig:vcg_with_alpha}
    }
    \caption{Influence of downsampling rate on MVBench(a) and VCG Score(b) performance. x-axis indicate the }
    \label{fig:ablate_alpha}
\end{figure}
Post-training optimization is defined as the combination of the LLMs' parameters of image MLLMs and learned LLMs' LoRA weights from video samples. A suitable fusion ratio could be highly efficient in boosting model performance trained under low-quality video-text samples. Here, we discuss the influence of different choices of fusion ratio on the understanding performance. As shown in ~\autoref{fig:ablate_alpha}, the x-axis represents the alpha value of LoRA. 0 indicates no LoRA weights added, and 32 means the LoRA weights are fully applied to LLMs. 
We observed distinct trends between MVBench and VCG Score. The former exhibits a peak around alpha 20, while the latter performs best near alpha 4.
This variance can be attributed to the nature of these two benchmarks: VCG typically involves longer length generations, whereas MVBench focuses on multiple-choice question answering, placing less emphasis on language generation ability. Consequently, weights learned from video-text data samples are more tailored for MVBench tasks. In this way, a larger portion of video weights are beneficial for MVBench. Moreover, from these two figures, it's evident that combining video and image weights leads to better performance than at the extremes of 0 and 32.

\subsection{Case Studies}
\begin{figure}[t]
    \centering
    \subfigure[Street Saxophone. ]{
    \begin{minipage}[b]{0.8\textwidth}
        \includegraphics[width=1\textwidth]{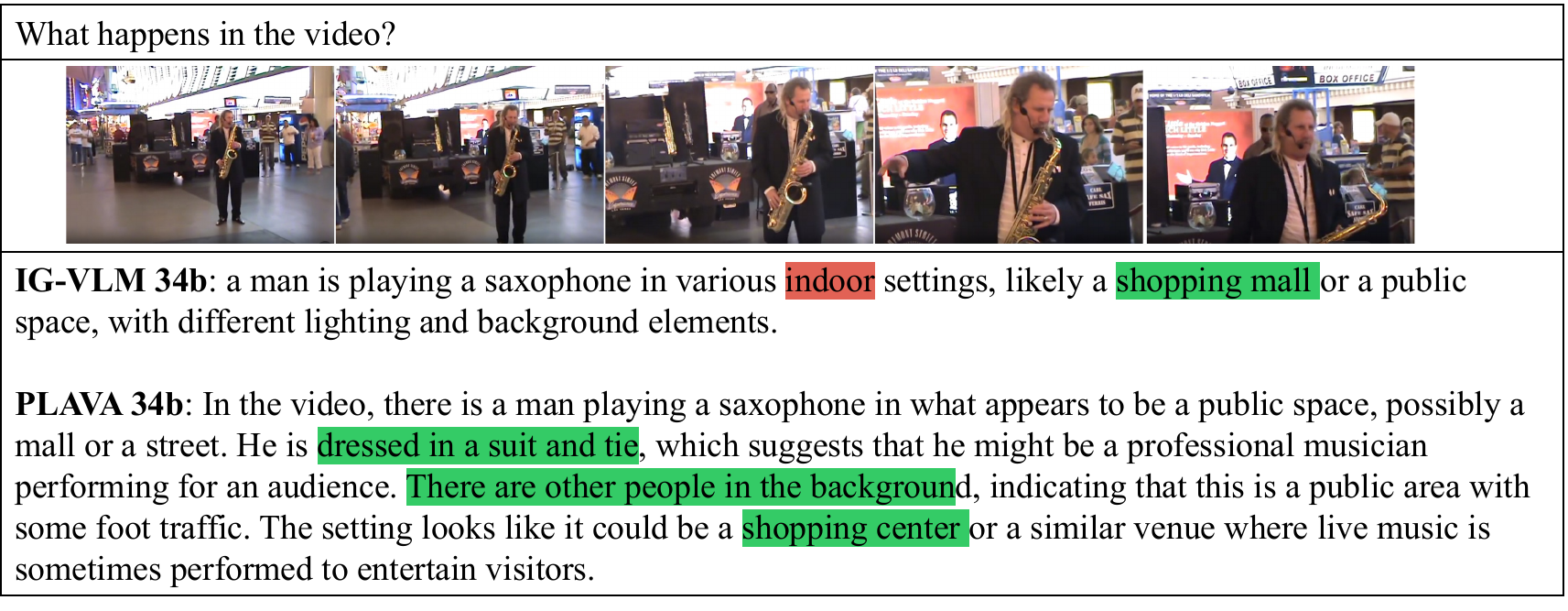}
    \end{minipage}
    \label{subfig:case_saxophone}
    }
    \subfigure[Badminton match.]{
    \begin{minipage}[b]{0.8\textwidth}
        \includegraphics[width=1\textwidth]{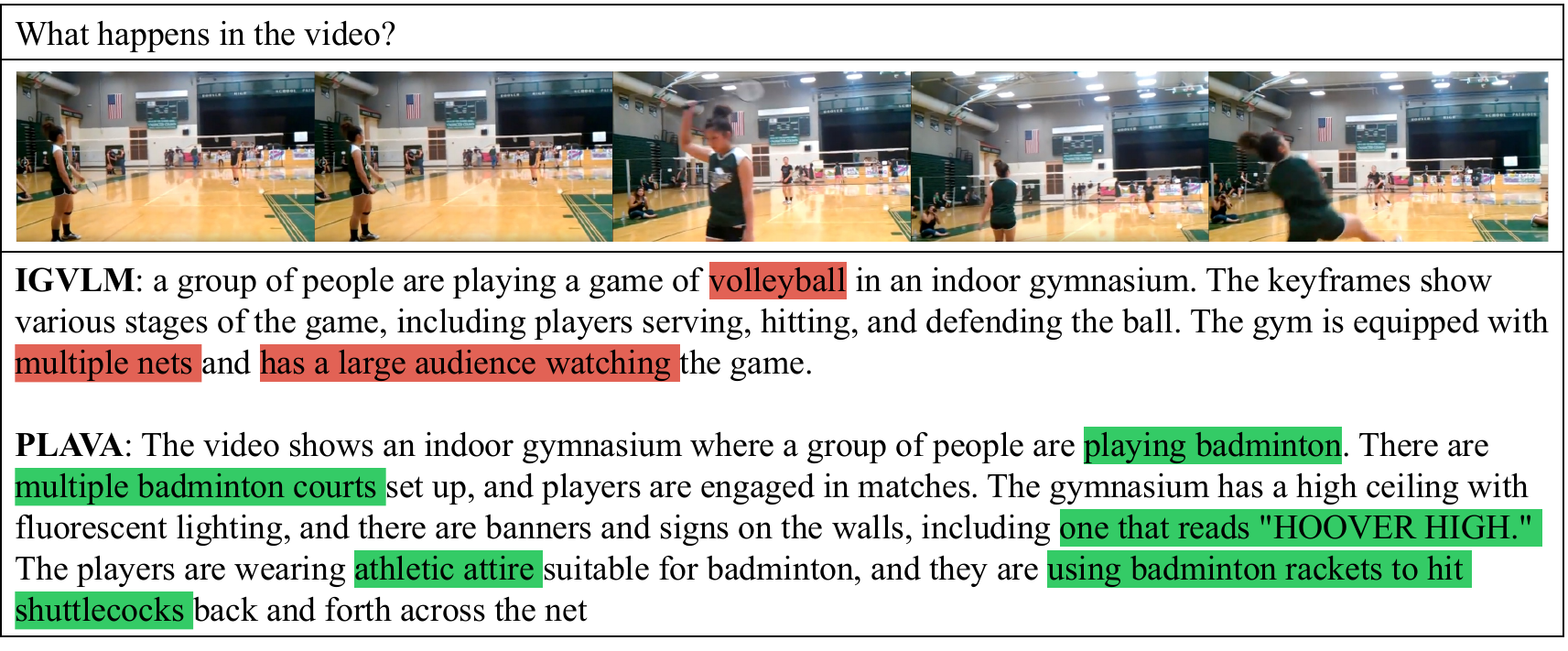}
    \end{minipage}
    \label{subfig:case_badminton}
    }

    \caption{Case Studies.}
    \label{fig:case_study}
\end{figure}

Apart from these quantitative results, we also qualitatively investigate the video understanding abilities of \nameofmethod models. We have shown several caption examples in~\autoref{fig:case_study}. According to the video clips, compared to IG-VLM, \nameofmethod 34B recognizes more details about videos, including the clothes worn by the main characters, the environment, and even some of the words in this video. Besides, as shown in Figure~\ref{subfig:case_badminton}, \nameofmethod can understand the video content more correctly, in which people are playing badminton rather than volleyball. These mistakes made by IG-VLM could be caused by the lowered resolution when concatenating frames into the grid view in the method design. Pooling reduces dimension after frames are encoded, thus leading to less information loss.

\subsection{Dense Recaption}
\begin{figure}[ht]
    \centering
    \includegraphics[width=0.8\textwidth]{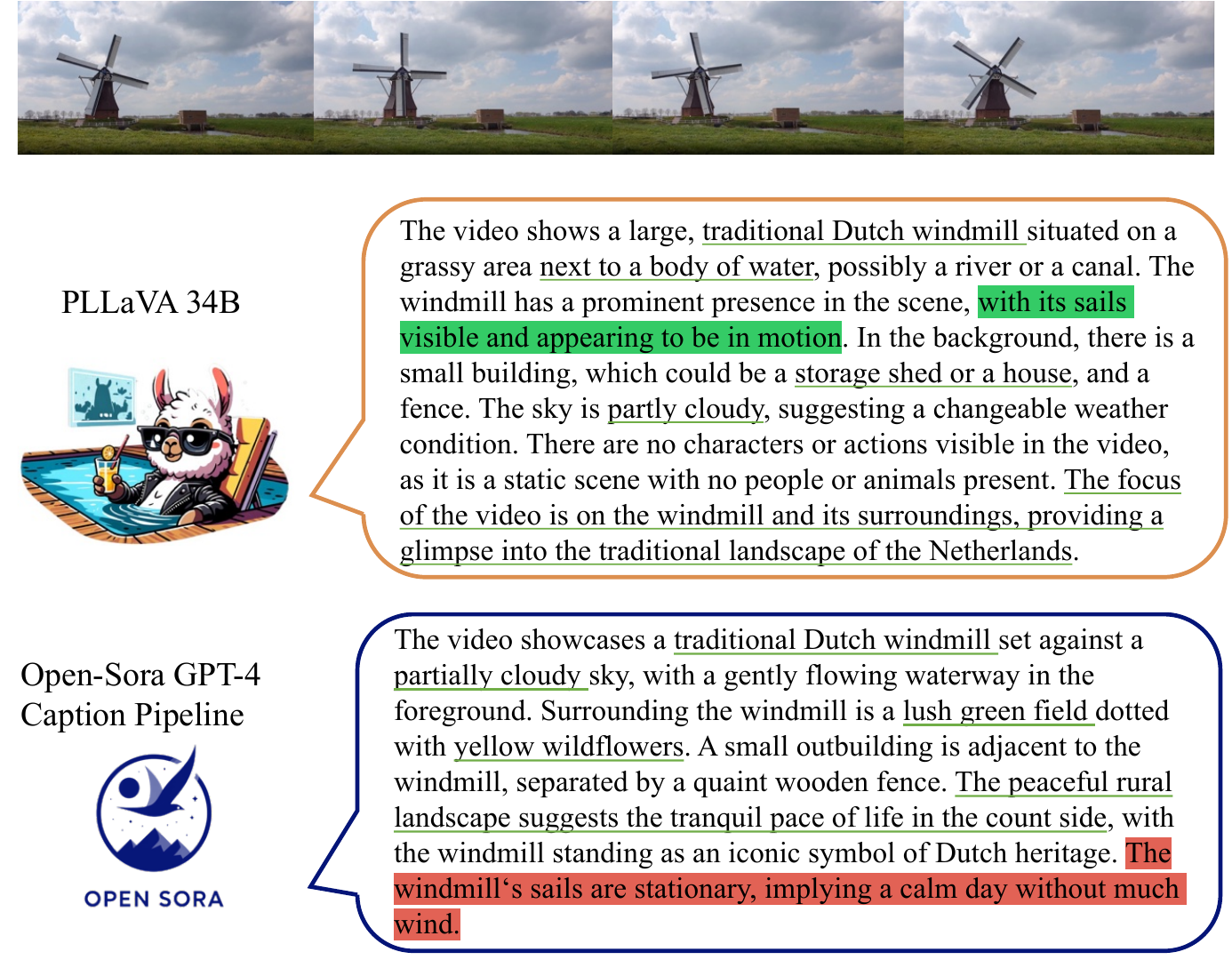}
    \caption{Recaption comparison between \nameofmethod 34B and Open-Sora.}
    \label{fig:recaption}
\end{figure}

In view of the caption ability of \nameofmethod, we further tested its recaption task and contributed 1K video Inter4K~\cite{stergiou2022adapool} caption dataset. An example is shown in ~\autoref{fig:recaption}. Compared to Open-Sora GPT-4 pipeline, our model captures better caption details and also highlights motion information in the video, demonstrate \nameofmethod's potential to contribute to the video generation community.
\section{Conclusion}

In this paper, we conduct an initial investigation for extending image-language models to videos with a simple yet extremely effective method, termed \nameofmethod. With the new model, it is easier to scale the training with more data and larger large language models with a more controllable strategy for over-training and performance saturation. \nameofmethod's ability of giving detailed captions also contributes to the community development of multimodal understanding and generation.

\clearpage
\bibliography{custom}

\begin{thebibliography}{10}

\bibitem{achiam2023gpt}
Josh Achiam, Steven Adler, Sandhini Agarwal, Lama Ahmad, Ilge Akkaya, Florencia~Leoni Aleman, Diogo Almeida, Janko Altenschmidt, Sam Altman, Shyamal Anadkat, et~al.
\newblock Gpt-4 technical report.
\newblock {\em arXiv preprint arXiv:2303.08774}, 2023.

\bibitem{bain2021webvid}
Max Bain, Arsha Nagrani, G{\"u}l Varol, and Andrew Zisserman.
\newblock Frozen in time: A joint video and image encoder for end-to-end retrieval.
\newblock In {\em Proceedings of the IEEE/CVF International Conference on Computer Vision}, pages 1728--1738, 2021.

\bibitem{chen2023videollm}
Guo Chen, Yin-Dong Zheng, Jiahao Wang, Jilan Xu, Yifei Huang, Junting Pan, Yi~Wang, Yali Wang, Yu~Qiao, Tong Lu, et~al.
\newblock Videollm: Modeling video sequence with large language models.
\newblock {\em arXiv preprint arXiv:2305.13292}, 2023.

\bibitem{chen2021evaluating}
Mark Chen, Jerry Tworek, Heewoo Jun, Qiming Yuan, Henrique Ponde de~Oliveira Pinto, Jared Kaplan, Harri Edwards, Yuri Burda, Nicholas Joseph, Greg Brockman, et~al.
\newblock Evaluating large language models trained on code.
\newblock {\em arXiv preprint arXiv:2107.03374}, 2021.

\bibitem{chiang2023vicuna}
Wei-Lin Chiang, Zhuohan Li, Zi~Lin, Ying Sheng, Zhanghao Wu, Hao Zhang, Lianmin Zheng, Siyuan Zhuang, Yonghao Zhuang, Joseph~E Gonzalez, et~al.
\newblock Vicuna: An open-source chatbot impressing gpt-4 with 90\%* chatgpt quality.
\newblock {\em See https://vicuna. lmsys. org (accessed 14 April 2023)}, 2(3):6, 2023.

\bibitem{Choudhury2023ZeroShotVQ}
Rohan Choudhury, Koichiro Niinuma, Kris~M. Kitani, and L{\'a}szl{\'o}~A. Jeni.
\newblock Zero-shot video question answering with procedural programs.
\newblock {\em ArXiv abs/2312.00937}, 2023.

\bibitem{feng2024mixture}
Wenfeng Feng, Chuzhan Hao, Yuewei Zhang, Yu~Han, and Hao Wang.
\newblock Mixture-of-loras: An efficient multitask tuning for large language models.
\newblock {\em arXiv preprint arXiv:2403.03432}, 2024.

\bibitem{goyal2017sthsthv2}
Raghav Goyal, Samira Ebrahimi~Kahou, Vincent Michalski, Joanna Materzynska, Susanne Westphal, Heuna Kim, Valentin Haenel, Ingo Fruend, Peter Yianilos, Moritz Mueller-Freitag, et~al.
\newblock The" something something" video database for learning and evaluating visual common sense.
\newblock In {\em Proceedings of the IEEE international conference on computer vision}, pages 5842--5850, 2017.

\bibitem{Grauman2021Ego4DAT}
Kristen Grauman, Andrew Westbury, Eugene Byrne, Zachary Chavis, Antonino Furnari, Rohit Girdhar, Jackson Hamburger, Hao Jiang, Miao Liu, and Xingyu~Liu et~al.
\newblock Ego4d: Around the world in 3,000 hours of egocentric video.
\newblock {\em IEEE Conf. Comput. Vis. Pattern Recog.}, pages 18995--19012, 2022.

\bibitem{hong2023cogagent}
Wenyi Hong, Weihan Wang, Qingsong Lv, Jiazheng Xu, Wenmeng Yu, Junhui Ji, Yan Wang, Zihan Wang, Yuxuan Zhang, Juanzi Li, Bin Xu, Yuxiao Dong, Ming Ding, and Jie Tang.
\newblock Cogagent: A visual language model for gui agents.
\newblock {\em ArXiv}, abs/2312.08914, 2023.

\bibitem{hu2021lora}
Edward~J Hu, Phillip Wallis, Zeyuan Allen-Zhu, Yuanzhi Li, Shean Wang, Lu~Wang, Weizhu Chen, et~al.
\newblock Lora: Low-rank adaptation of large language models.
\newblock In {\em International Conference on Learning Representations}, 2021.

\bibitem{huang2023vtimellm}
Bin Huang, Xin Wang, Hong Chen, Zihan Song, and Wenwu Zhu.
\newblock Vtimellm: Empower llm to grasp video moments, 2023.

\bibitem{huang2024lita}
De-An Huang, Shijia Liao, Subhashree Radhakrishnan, Hongxu Yin, Pavlo Molchanov, Zhiding Yu, and Jan Kautz.
\newblock Lita: Language instructed temporal-localization assistant.
\newblock {\em arXiv preprint arXiv:2403.19046}, 2024.

\bibitem{Jin2023ChatUniViUV}
Peng Jin, Ryuichi Takanobu, Caiwan Zhang, Xiaochun Cao, and Li~Yuan.
\newblock Chat-univi: Unified visual representation empowers large language models with image and video understanding.
\newblock {\em ArXiv abs/2311.08046}, 2024.

\bibitem{kay2017kinetics}
Will Kay, Joao Carreira, Karen Simonyan, Brian Zhang, Chloe Hillier, Sudheendra Vijayanarasimhan, Fabio Viola, Tim Green, Trevor Back, Paul Natsev, et~al.
\newblock The kinetics human action video dataset.
\newblock {\em arXiv preprint arXiv:1705.06950}, 2017.

\bibitem{kim2024igvlm}
Wonkyun Kim, Changin Choi, Wonseok Lee, and Wonjong Rhee.
\newblock An image grid can be worth a video: Zero-shot video question answering using a vlm.
\newblock {\em arXiv preprint arXiv:2403.18406}, 2024.

\bibitem{lecun1989handwritten}
Yann LeCun, Bernhard Boser, John Denker, Donnie Henderson, Richard Howard, Wayne Hubbard, and Lawrence Jackel.
\newblock Handwritten digit recognition with a back-propagation network.
\newblock {\em Advances in neural information processing systems}, 2, 1989.

\bibitem{li2023blip2}
Junnan Li, Dongxu Li, Silvio Savarese, and Steven Hoi.
\newblock Blip-2: Bootstrapping language-image pre-training with frozen image encoders and large language models.
\newblock In {\em International conference on machine learning}, pages 19730--19742. PMLR, 2023.

\bibitem{li2023videochat}
KunChang Li, Yinan He, Yi~Wang, Yizhuo Li, Wenhai Wang, Ping Luo, Yali Wang, Limin Wang, and Yu~Qiao.
\newblock Videochat: Chat-centric video understanding.
\newblock {\em arXiv preprint arXiv:2305.06355}, 2023.

\bibitem{li2023mvbench}
Kunchang Li, Yali Wang, Yinan He, Yizhuo Li, Yi~Wang, Yi~Liu, Zun Wang, Jilan Xu, Guo Chen, Ping Luo, et~al.
\newblock Mvbench: A comprehensive multi-modal video understanding benchmark.
\newblock {\em ArXiv abs/2311.17005}, 2023.

\bibitem{Li2023LLaMAVIDAI}
Yanwei Li, Chengyao Wang, and Jiaya Jia.
\newblock Llama-vid: An image is worth 2 tokens in large language models.
\newblock {\em ArXiv abs/2311.17043}, 2023.

\bibitem{li2016tgif}
Yuncheng Li, Yale Song, Liangliang Cao, Joel Tetreault, Larry Goldberg, Alejandro Jaimes, and Jiebo Luo.
\newblock Tgif: A new dataset and benchmark on animated gif description.
\newblock In {\em Proceedings of the IEEE Conference on Computer Vision and Pattern Recognition}, pages 4641--4650, 2016.

\bibitem{lian2022scaling}
Dongze Lian, Daquan Zhou, Jiashi Feng, and Xinchao Wang.
\newblock Scaling \& shifting your features: A new baseline for efficient model tuning.
\newblock {\em Advances in Neural Information Processing Systems}, 35:109--123, 2022.

\bibitem{lian2023llm}
Long Lian, Boyi Li, Adam Yala, and Trevor Darrell.
\newblock Llm-grounded diffusion: Enhancing prompt understanding of text-to-image diffusion models with large language models.
\newblock {\em arXiv preprint arXiv:2305.13655}, 2023.

\bibitem{Lin2023VideoLLaVALU}
Bin Lin, Bin Zhu, Yang Ye, Munan Ning, Peng Jin, and Li~Yuan.
\newblock Video-llava: Learning united visual representation by alignment before projection.
\newblock {\em ArXiv abs/2311.10122}, 2023.

\bibitem{lin2023vila}
Ji~Lin, Hongxu Yin, Wei Ping, Yao Lu, Pavlo Molchanov, Andrew Tao, Huizi Mao, Jan Kautz, Mohammad Shoeybi, and Song Han.
\newblock Vila: On pre-training for visual language models.
\newblock {\em arXiv preprint arXiv:2312.07533}, 2023.

\bibitem{liu2023improved}
Haotian Liu, Chunyuan Li, Yuheng Li, and Yong~Jae Lee.
\newblock Improved baselines with visual instruction tuning.
\newblock In {\em NeurIPS 2023 Workshop on Instruction Tuning and Instruction Following}, 2023.

\bibitem{liu2024llavanext}
Haotian Liu, Chunyuan Li, Yuheng Li, Bo~Li, Yuanhan Zhang, Sheng Shen, and Yong~Jae Lee.
\newblock Llava-next: Improved reasoning, ocr, and world knowledge, January 2024.

\bibitem{liu2024visual}
Haotian Liu, Chunyuan Li, Qingyang Wu, and Yong~Jae Lee.
\newblock Visual instruction tuning.
\newblock {\em Advances in neural information processing systems}, 36, 2024.

\bibitem{liu2023btadapter}
Ruyang Liu, Chen Li, Yixiao Ge, Ying Shan, Thomas~H Li, and Ge~Li.
\newblock One for all: Video conversation is feasible without video instruction tuning.
\newblock {\em arXiv preprint arXiv:2309.15785}, 2023.

\bibitem{liu2024st}
Ruyang Liu, Chen Li, Haoran Tang, Yixiao Ge, Ying Shan, and Ge~Li.
\newblock St-llm: Large language models are effective temporal learners.
\newblock {\em arXiv preprint arXiv:2404.00308}, 2024.

\bibitem{Ma2023VistaLLaMARV}
Fan Ma, Xiaojie Jin, Heng Wang, Yuchen Xian, Jiashi Feng, and Yi~Yang.
\newblock Vista-llama: Reliable video narrator via equal distance to visual tokens.
\newblock {\em ArXiv abs/2312.08870}, 2023.

\bibitem{maaz2023videochatgpt}
Muhammad Maaz, Hanoona Rasheed, Salman Khan, and Fahad~Shahbaz Khan.
\newblock Video-chatgpt: Towards detailed video understanding via large vision and language models.
\newblock {\em arXiv preprint arXiv:2306.05424}, 2023.

\bibitem{chatgpt}
OpenAI.
\newblock Chatgpt.
\newblock \url{https://openai.com/blog/chatgpt}, 2023.

\bibitem{radford2021learning}
Alec Radford, Jong~Wook Kim, Chris Hallacy, Aditya Ramesh, Gabriel Goh, Sandhini Agarwal, Girish Sastry, Amanda Askell, Pamela Mishkin, Jack Clark, et~al.
\newblock Learning transferable visual models from natural language supervision.
\newblock In {\em International conference on machine learning}, pages 8748--8763. PMLR, 2021.

\bibitem{roumeliotis2023llama}
Konstantinos~I Roumeliotis, Nikolaos~D Tselikas, and Dimitrios~K Nasiopoulos.
\newblock Llama 2: Early adopters' utilization of meta's new open-source pretrained model.
\newblock 2023.

\bibitem{Song2023MovieChatFD}
Enxin Song, Wenhao Chai, Guanhong Wang, Yucheng Zhang, Haoyang Zhou, Feiyang Wu, Xun Guo, Tianbo Ye, Yang Lu, Jenq-Neng Hwang, and Gaoang Wang.
\newblock Moviechat: From dense token to sparse memory for long video understanding.
\newblock {\em ArXiv abs/2307.16449}, 2023.

\bibitem{stergiou2022adapool}
Alexandros Stergiou and Ronald Poppe.
\newblock Adapool: Exponential adaptive pooling for information-retaining downsampling.
\newblock {\em IEEE Transactions on Image Processing}, 32:251--266, 2022.

\bibitem{suris2023vipergpt}
D\'idac Sur\'is, Sachit Menon, and Carl Vondrick.
\newblock Vipergpt: Visual inference via python execution for reasoning.
\newblock {\em Proceedings of IEEE International Conference on Computer Vision (ICCV)}, 2023.

\bibitem{touvron2023llama}
Hugo Touvron, Thibaut Lavril, Gautier Izacard, Xavier Martinet, Marie-Anne Lachaux, Timoth{\'e}e Lacroix, Baptiste Rozi{\`e}re, Naman Goyal, Eric Hambro, Faisal Azhar, et~al.
\newblock Llama: Open and efficient foundation language models.
\newblock {\em arXiv preprint arXiv:2302.13971}, 2023.

\bibitem{vaswani2017attention}
Ashish Vaswani, Noam Shazeer, Niki Parmar, Jakob Uszkoreit, Llion Jones, Aidan~N Gomez, {\L}ukasz Kaiser, and Illia Polosukhin.
\newblock Attention is all you need.
\newblock {\em Advances in neural information processing systems}, 30, 2017.

\bibitem{wu2023textvr}
Weijia Wu, Yuzhong Zhao, Zhuang Li, Jiahong Li, Hong Zhou, Mike~Zheng Shou, and Xiang Bai.
\newblock A large cross-modal video retrieval dataset with reading comprehension.
\newblock {\em arXiv preprint arXiv:2305.03347}, 2023.

\bibitem{xiao2021nextqa}
Junbin Xiao, Xindi Shang, Angela Yao, and Tat-Seng Chua.
\newblock Next-qa: Next phase of question-answering to explaining temporal actions.
\newblock In {\em Proceedings of the IEEE/CVF conference on computer vision and pattern recognition}, pages 9777--9786, 2021.

\bibitem{xu2017videoqa}
Dejing Xu, Zhou Zhao, Jun Xiao, Fei Wu, Hanwang Zhang, Xiangnan He, and Yueting Zhuang.
\newblock Video question answering via gradually refined attention over appearance and motion.
\newblock In {\em Proceedings of the 25th ACM international conference on Multimedia}, pages 1645--1653, 2017.

\bibitem{Yang2022FrozenBiLM}
Antoine Yang, Antoine Miech, Josef Sivic, Ivan Laptev, and Cordelia Schmid.
\newblock Zero-shot video question answering via frozen bidirectional language models.
\newblock {\em Adv. Neural Inform. Process. Syst.}, 35:124--141, 2022.

\bibitem{ye2024cat}
Qilang Ye, Zitong Yu, Rui Shao, Xinyu Xie, Philip Torr, and Xiaochun Cao.
\newblock Cat: Enhancing multimodal large language model to answer questions in dynamic audio-visual scenarios.
\newblock {\em arXiv preprint arXiv:2403.04640}, 2024.

\bibitem{yi2020clevrer}
Kexin Yi, Chuang Gan, Yunzhu Li, Pushmeet Kohli, Jiajun Wu, Antonio Torralba, and Joshua~B Tenenbaum.
\newblock Clevrer: Collision events for video representation and reasoning.
\newblock In {\em International Conference on Learning Representations}, 2020.

\bibitem{yu2019activitynet}
Zhou Yu, Dejing Xu, Jun Yu, Ting Yu, Zhou Zhao, Yueting Zhuang, and Dacheng Tao.
\newblock Activitynet-qa: A dataset for understanding complex web videos via question answering.
\newblock In {\em AAAI}, pages 9127--9134, 2019.

\bibitem{Zhang2023ASL}
Ce~Zhang, Taixi Lu, Md~Mohaiminul Islam, Ziyang Wang, Shoubin Yu, Mohit Bansal, and Gedas Bertasius.
\newblock A simple llm framework for long-range video question-answering.
\newblock {\em ArXiv abs/2312.17235}, 2023.

\bibitem{zhang2023VideoLLAMA}
Hang Zhang, Xin Li, and Lidong Bing.
\newblock Video-{LL}a{MA}: An instruction-tuned audio-visual language model for video understanding.
\newblock In {\em Conf. Empirical Methods in Natural Language Processing}, pages 543--553, 2023.

\bibitem{Zhang2023LLaMAAdapterEF}
Renrui Zhang, Jiaming Han, Aojun Zhou, Xiangfei Hu, Shilin Yan, Pan Lu, Hongsheng Li, Peng Gao, and Yu~Qiao.
\newblock Llama-adapter: Efficient fine-tuning of language models with zero-init attention.
\newblock {\em arXiv preprint arXiv:2303.16199}, 2023.

\bibitem{zhou2018youcook2}
Luowei Zhou, Chenliang Xu, and Jason Corso.
\newblock Towards automatic learning of procedures from web instructional videos.
\newblock In {\em Proceedings of the AAAI Conference on Artificial Intelligence}, 2018.

\bibitem{zhu2023minigpt}
Deyao Zhu, Jun Chen, Xiaoqian Shen, Xiang Li, and Mohamed Elhoseiny.
\newblock Minigpt-4: Enhancing vision-language understanding with advanced large language models.
\newblock In {\em The Twelfth International Conference on Learning Representations}, 2023.

\end{thebibliography}
\bibliographystyle{plain}

\end{document}